# Parallel Coordinates for Discovery of Interpretable Machine Learning Models

Dustin Hayes, Boris Kovalerchuk
Dept. of Computer Science, Central Washington University, USA.
Dustin.Hayes@cwu.edu, BorisK@cwu.edu

*Abstract*—This work uses visual knowledge discovery in parallel coordinates to advance methods of interpretable machine learning. The graphic data representation in parallel coordinates made the concepts of hypercubes and hyperblocks (HBs) simple to understand for end users. It is suggested to use mixed and pure hyperblocks in the proposed data classifier algorithm Hyper. It is shown that Hyper models generalize decision trees. The algorithm is presented in several settings and options to discover interactively or automatically overlapping or non-overlapping hyperblocks. Additionally, the use of hyperblocks in conjunction with language descriptions of visual patterns is demonstrated. The benchmark data from the UCI ML repository were used to evaluate the Hyper algorithm. It enabled the discovery of mixed and pure HBs evaluated using 10-fold cross validation. Connections among hyperblocks, dimension reduction and visualization have been established. The capability of end users to find and observe hyperblocks, as well as the ability of side-by-side visualizations to make patterns evident, are among major advantages of hyperblock technology and the Hyper algorithm. A new method to visualize incomplete n-D data with missing values is proposed, while the traditional parallel coordinates do not support it. The ability of HBs to better prevent both overgeneralization and overfitting of data over decision trees is demonstrated as another benefit of the hyperblocks. The features of VisCanvas 2.0 software tool that implements Hyper technology are presented.

## 1. Introduction

Acceptance, interpretability, and comprehensibility of different classifiers are crucial for future Machine Learning (ML) advancements. For many machine learning models this is a very significant challenge due to their black box specifics making these models incomprehensible. Users are reluctant to deploy such models for high-risk, high-stakes decisions. A promising solution to this problem is visual knowledge discovery [5-7, 15, 18, 22]. We outline a parallel coordinates-based visual knowledge discovery approach below that includes supervised learning, data and model visualization, dimensionality reduction, and model simplification.

The supervised classification models are the main topic of this work. Often, developing a **reliable interpretable, explainable, and comprehensible ML model** necessitates placing the end-user in control of the creation of a model. The end users are frequently **subject matter experts** rather than machine learning professionals. Often for them formal ML models are opaque black boxes. The **visual knowledge discovery** (VKD) methodology makes it possible to identify ML models, explain them for the end users, and put them in charge of model development. This work extends [15] in: (1) visualization method for data with empty values and more elaborated examples, (2) dealing with the large data, (3) in presenting the features of software tool developed to support visual knowledge discovery in parallel coordinates denoted as VisCanvas 2.0, and (4) generalization of hyperblock approach to other general line coordinates.

The suitability of parallel coordinates for visual knowledge discovery is demonstrated in many prior works [1-3, 10, 13, 19-21]. Parallel coordinates accomplish so without losing any of the multidimensional information, and they support interpretability and comprehensibility by using the original attributes, which have clear domain meaning for the domain's end users. However, the use of parallel coordinates for supervised learning as a primary space for actual solving supervised learning tasks still is in the nascent stage.

The suggested approach is based on the ideas of hyperblocks and hypercubes, which generalize 2-D squares and rectangles to n-D space. They are naturally represented in parallel coordinates and have a distinct interpretation. It gives the end users the chance to actively participate in developing and enhancing these models in addition to joust using models. It has been suggested to develop decision tree models with parallel coordinates in [1, 11]. We demonstrate that the hyperblock technique produces **more general models** than decision trees (DTs). It allows us to deal better with both overfitting and overgeneralization of data.

The **major contributions** of this work are in:

- employing *hypercubes* and *hyperblocks* to assist users discover patterns in parallel coordinates for the supervised learning tasks,
- combining *hyperblocks, visualizations,* and *language descriptions* to discover and present patterns,
- allowing *users to lead* the processes of development, enhancement, and usage of the models interactively,
- handling missing values in large datasets, which overcomes a weakness of traditional parallel coordinates.

The chapter is organized as follows. Section 2 defines (1) hypercubes and hyperblocks, together with their parallel coordinate visualization, and (2) algorithms to find them automatically and interactively. We first concentrate on individual HCs and HBs, then on multiple HBs. This includes finding and visualizing pairs of non-overlapping HBs and HB combinations along with linguistic descriptions of visual patterns.

Supervised learning in parallel coordinates is presented in Section 3. We first discuss its difficulties before introducing the Hyper classification technique for pure and mixed/impure hyperblocks. Following the comparison with decision trees, hyper models are presented as generalized decision trees. The case study uses benchmark Wisconsin Breast Cancer (WBC) data from UCI ML repository to test the methodology.

To accomplish this, we first provide learning of hyperblocks using all WBC data, followed by 10-fold cross validation using training and validation data. The relationship between hyperblocks, dimension reduction, and the visualization of lower dimensional hyperblocks has been established.

Section 4 presents the features of VisCanvas 2.0 software [23], which implements Hyper methodology and is based on the prior VisCanvas 1.0 [12]. Both are available on GitHub [12,23], This description includes the method of visualizing cases with missing values. Section 5 covers discussion, generalization, conclusion, and future work.

## 2. Interactive and automatic visual discovery of hyperblocks

### *2.1. Hypercubes and hyperblocks*

Interpretable concepts that appear naturally in parallel coordinates are hypercubes (HCs) and hyperblocks (HBs).

A **hypercube** (**n-cube**) is a set of n-D points $\{\mathbf{x}=(x_1,x_2,\ldots,x_n)\}$ with a **center** in n-D point $\mathbf{c}=(c_1,c_2,\ldots,c_n)$ and **side length** $L$ such that

$$\forall i \ \ \| x_i - c_i \| \le L/2 \qquad (1)$$

A hypercube is an n-dimensional generalization of ordinary square ($n$=2) and cube ($n$=3). A *unit hypercube* has $L$=1. A *binary hypercube* is an n-cube in the binary n-D space $E^n$.

A **hyperblock** (**hyperrectangle, n-orthotope**) is a set of n-D points $\{\mathbf{x}=(x_1,x_2,\ldots,x_n)\}$ with a **center** in n-D point $\mathbf{c}=(c_1,c_2,\ldots,c_n)$ and **side lengths** $\boldsymbol{L}=(L_1, L_2,\ldots, L_n)$ such that

$$\forall i \ \ \| x_i - c_i \| \le L_i/2 \qquad (2)$$

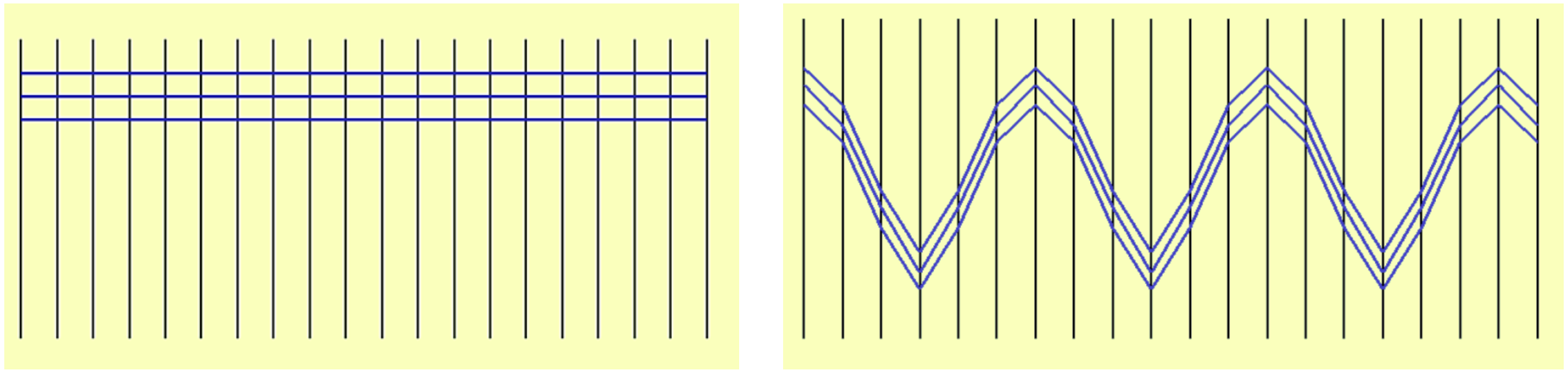

(a) hypercube A with equal coordinates (b) zigzag hypercube B

Fig. 1. Examples of hypercubes.

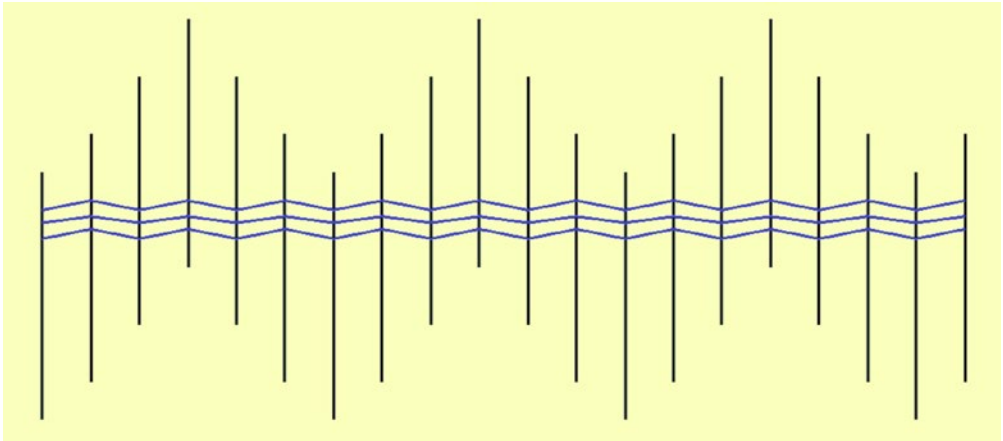

Fig. 2. Hypercubes in adjustable parallel coordinates shifted up and down to simplify hypercube patterns.

As a result, the *n*-cube is a unique instance of a hyperblock. Figs. 1 and 2 depict the idea of a hypercube. The hypercubes in these figures are sized in relation to one another in 20-D space based on their length. Compared to Fig. 1bc, the HCs in Fig. 1a are bigger. The central 20-D points of these HCs are shown by the middle lines.

Figure 2 shows hypercube patterns in adjustable parallel coordinates (APC), where coordinates can be sh Step M3.1ifted up and down, to make them easier to perceive [7]. Fig.3 shows all Wisconsin Breast Cancer (WBC) data in parallel coordinates. In this study we explore WBC data and build multiple hyperblock on them to discover an interpretable supervised model classifying benign and malignant cases.

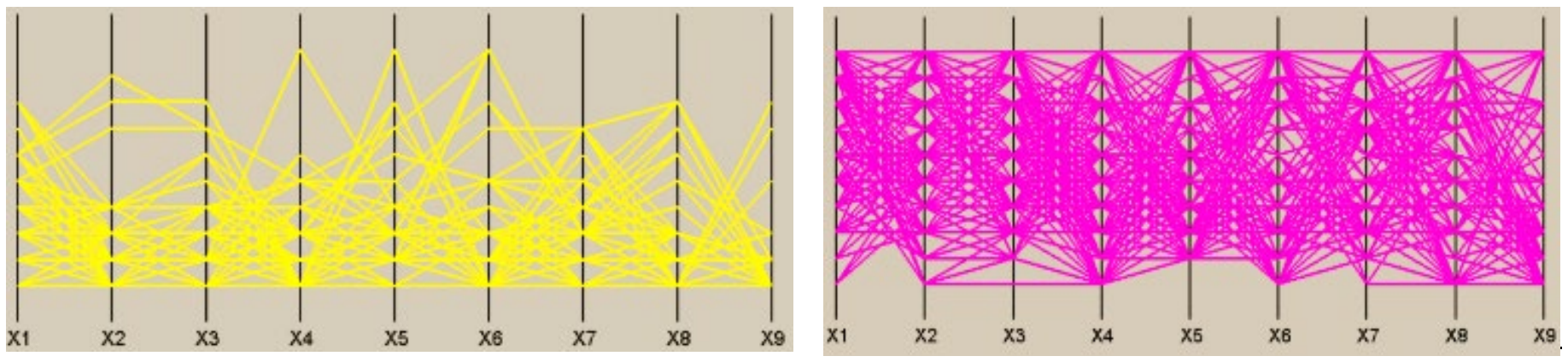

Fig 3. All cases of WBC data (yellow - benign, magenta- malignant) in parallel coordinates.

## *2.2. Algorithm to generate hyperblocks*

Hyperblocks can be made up of hypercubes whose centers are close to one another and share some dimensions. Not all HC pairings, nevertheless, will result in hyperblocks.

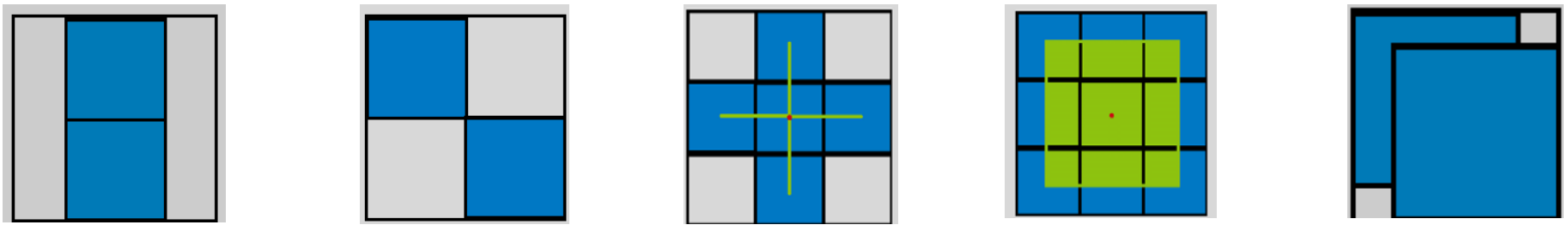

(a) (b) (c) (d) (e)

Fig. 4. Adjacency options for 2-D hypercubes.

Fig. 4 shows different alternatives how HCs can be combined. Fig. 4a shows two adjacent "2-D hypercubes" (squares) that share *n*-1 dimensions (1-D line for the squares) that can form a single hyperblock. Fig 4b shows two adjacent "2-D hypercubes" (squares) that share *n*-2 dimensions (a single point for the squares) that do not form a single hyperblock. Figs. 4cd show more n-1 and n-2 adjacent HCs for the square in the center, while Fig. 4e shows overlapping HCs that do not from a single hyperblock. Due to these complex relations between HCs and HBs, in this work, we start with small pure HBs seeded in the individual n-D points as centers and grow them to HBs keeping their purity as the steps below describe.

Steps of the **Merger Hyperblocks** (**MHyper**) algorithm for pure and dominant hyperblocks.

Step M1: Seed an initial set of **pure** HBs with a single n-D point in each of them (HBs with length equal to 0).

Step M2: Select a HB **x** from the set of all HBs.

Step M3: Start iterating over the remaining HBs. If $HB_i$ has the same class as **x** then attempt to combine $HB_i$ with HB **x** to get a pure HB.

- Step M3.1: Create a joint HB from $HB_i$ and HB **x** that is an envelope around $HB_i$ and HB **x** using the minimum and maximum of each dimension for $HB_i$ and HB **x.**
- Step M3.2: Check if any other n-D point **y** belongs to the envelop of $HB_i$ and HB **x.** If **y** belongs to this envelope add **y** to the joint HB.
- Step M3.3: If all points **y** in the joint HB are of the same class then remove **x** and $HB_i$ from the set of HBs that need to be changed.

Step M4: Repeat step 3 for all remaining HBs that need to be changed. The result is a *full pure HB* that cannot be extended with other n-D points and continue to be pure,

Step M5: Repeat step 2 for n-D points do not belong to already constructed full pure HBs.

Step M6: Define an *impurity threshold* that limits the percentage of n-D points from opposite classes allowed in a **dominant** HB.

Step M7: Select a HB **x** from the set of all HBs.

Step M8: Attempt to combine **x** with remaining HBs.

- Step M8.1. Create a joint HB from $HB_i$ and **x** that is an envelope of $HB_i$ and **x**.
- Step M8.2: Check if any other n-D point **y** belongs to the envelop of $HB_i$ and **x.** If **y** belongs to this envelope add **y** to the joint HB.
- Step M8.3: Compute impurity of the $HB_i$ (the percentage of n-D points from opposite classes introduced by the combination of **x** with $HB_i$.)
- Step M8.3 Find $HB_i$ with lowest impurity. If this lowest impurity is below predefined impurity threshold create a joint HB.

Step M9: Repeat step 7 until all combinations are made.

### *2.3. Individual hyperblocks*

The **overlapping** of lines (polylines) that represent individual cases (n-D points) is a significant **problem** for finding patterns, as was seen above. Even when the values are different, hundreds of lines can follow the same course and be displayed on top of one another in a given visualization resolution. Those lines are indistinguishable to the user. The user is therefore unaware of the number of lines in the hypercube and which line segments correspond to which n-D locations.

**Histograms for individual hyperblocks**. This problem is addressed in part by histograms. A distribution of n-D points (pink lines) within a hyperblock is shown in Fig. 5 as white rectangles and identical-length black lines. A data quantile is represented by the height of each white rectangle and black separation lines. The pink lines' widths represent their frequencies. This image demonstrates the distribution of these 9-D points in the 9-D WBC space.

The visual representation of the distribution of polylines in each hyperblock presented **side-by-side** with the display of quartiles in a separate window is shown in Figs. 6 and 7. The distribution of 9-D points in HBs A and B is shown in Figs. 6 and 7. It also make it possible to distinguish between them clearly, particularly in coordinate X6 where their distributions do not overlap while both HBs A and B from Figs. 6 and 7 belong to the same class.

In Fig 6. the white HB A contains pink and yellow lines from opposing classes with other HBs visible only partially on the background. Several yellow lines and a single pink line from the opposite class are used to depict another HB B in Fig. 7.

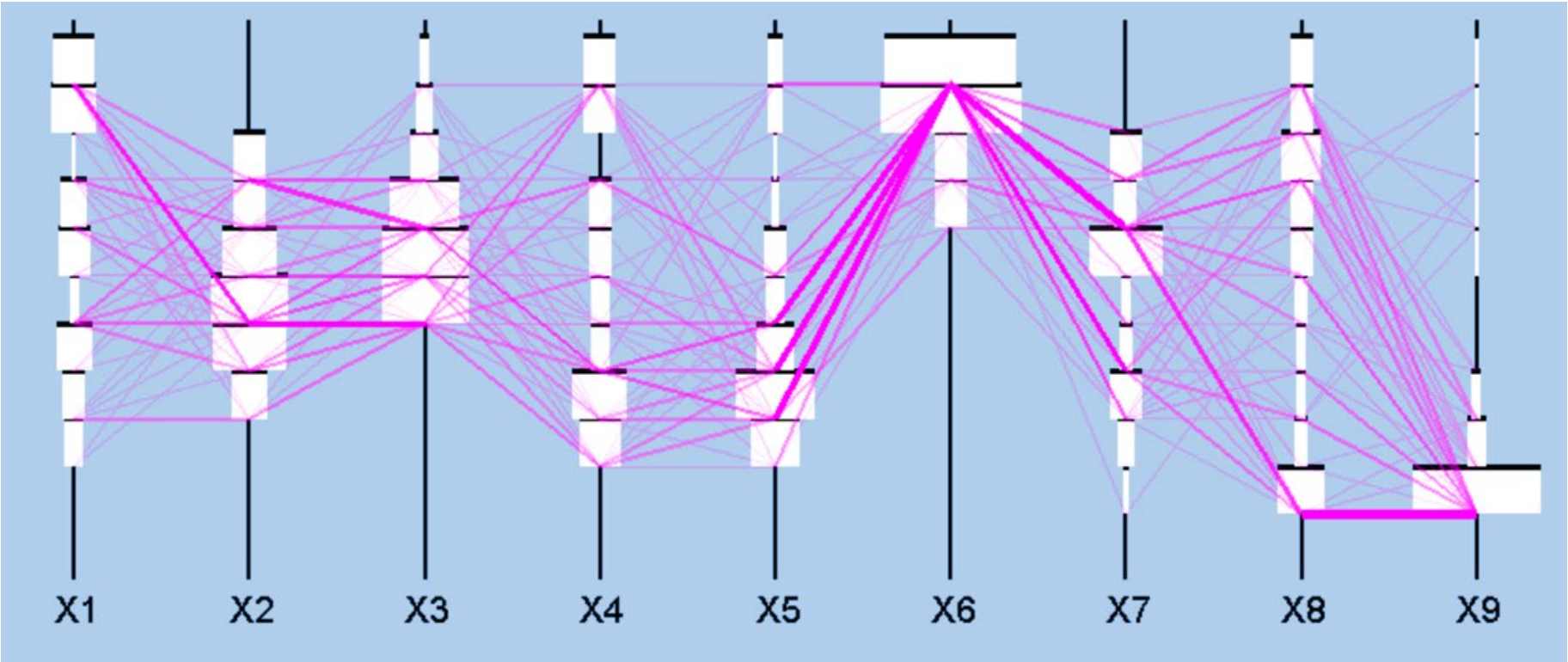

Fig. 5. Distribution of 9-D points with white rectangles to show quantiles for two WBC HBs.

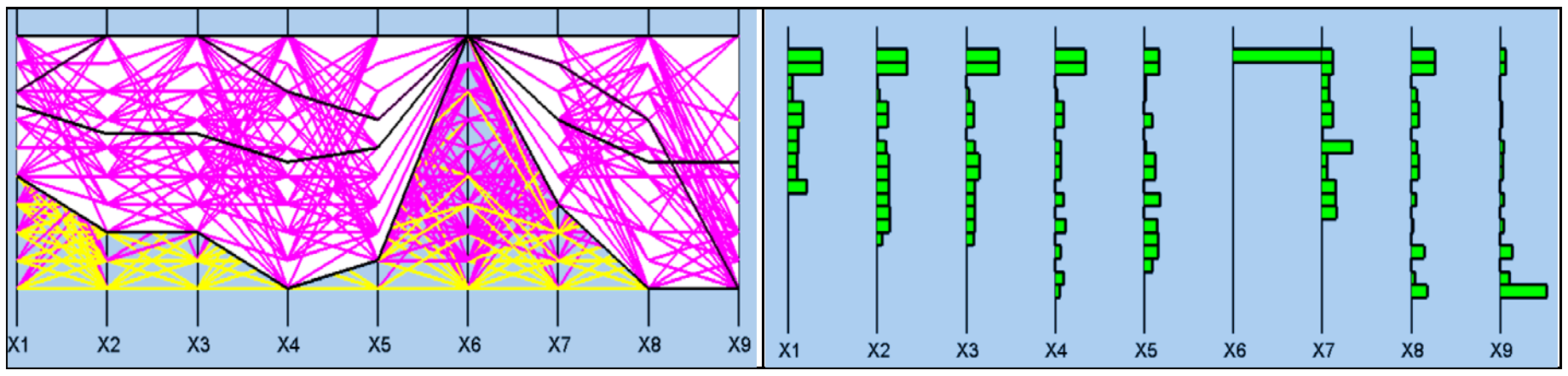

(a) HB A with white background (b) Frequency of cases for HB A.

Fig. 6: Hyperblock A and its distribution using quantiles.

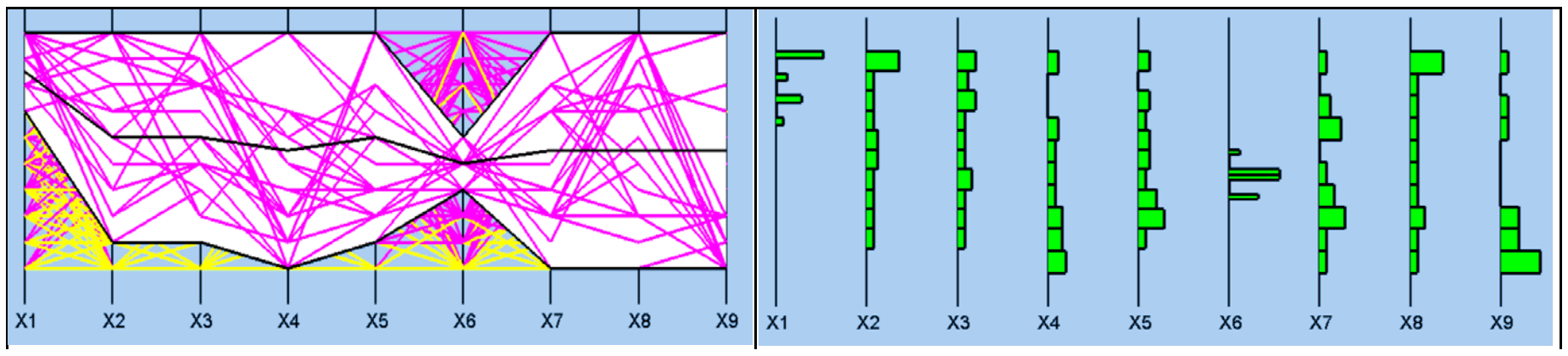

(a) HB B with white background (b) Frequency of cases for this hyperblock

Fig. 7: Hyperblock B and its distribution using quantiles.

In Fig. 8a, an illustration of well-separated HBs from opposing pink and yellow classes is shown. Their frequencies are presented by the width of the lines. As a result, we have embedded frequencies in these parallel coordinates.

These illustrations demonstrate that if we can find such different HBs, we can develop understandable categorization rules. Further visualization examples in parallel coordinates with embedded frequency by line width are shown in Figs. 8–10.

The selected hyperblock is displayed with a white background and a black line indicates the hyperblock's mean. A pure HB example is displayed in Fig. 10, whereas mixed hyperblock examples are shown in Fig. 11, where cases from various classes are depicted in various colors.

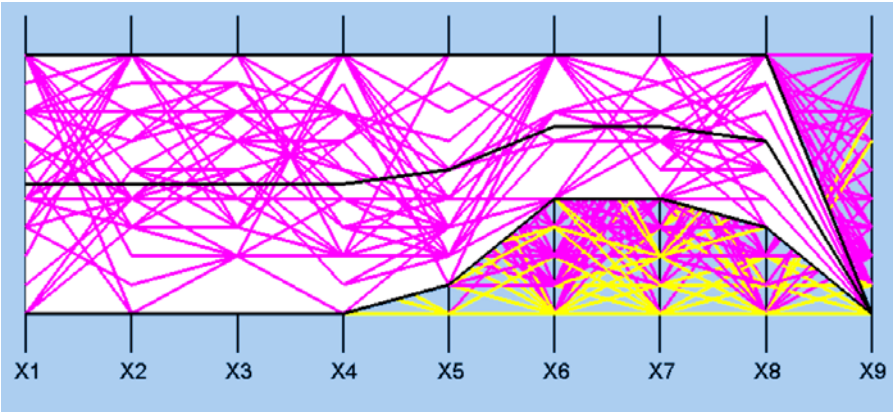

Fig. 8. Normal visualization of a hyperblock.

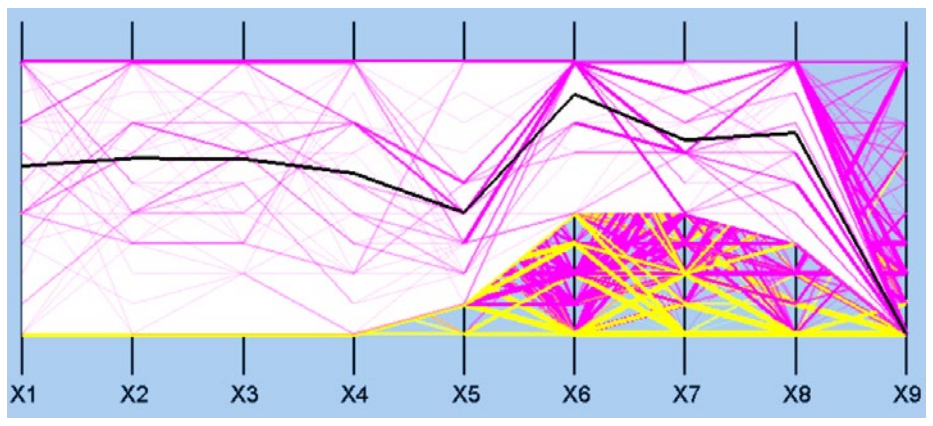

Fig. 9. A hyperblock with line frequencies enbeded.

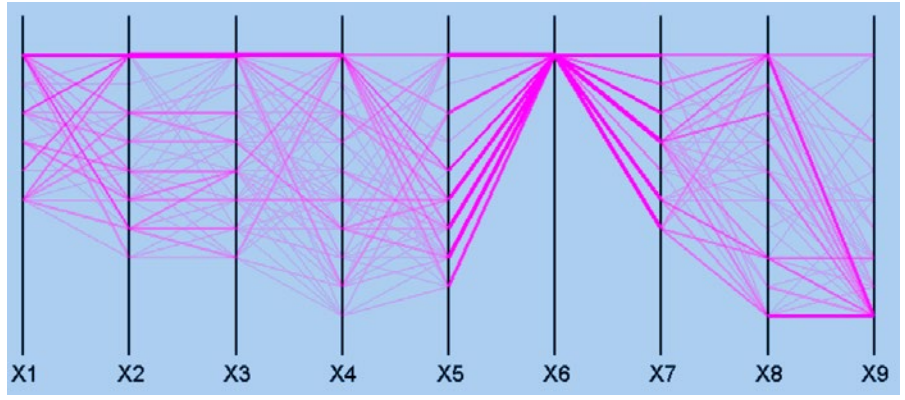

Fig. 10. Hyperblocks` lines with their frequencies

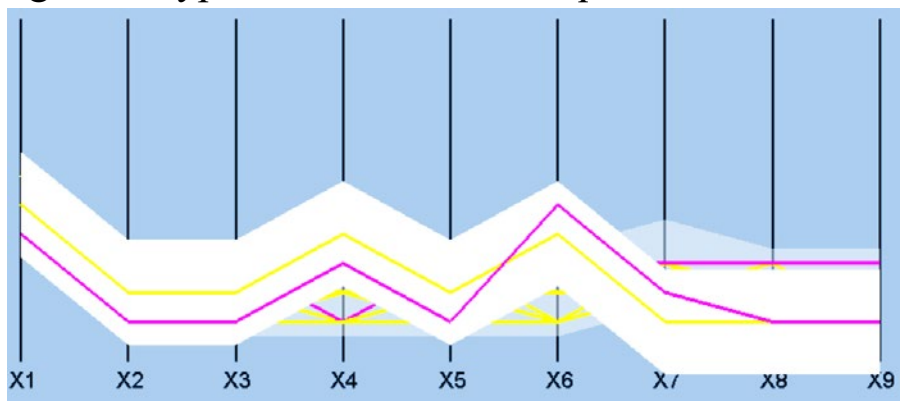

Fig. 11. Mixed hyperblocks.

### *2.4. Multiple hyperblocks*

**Hyperblocks viewed side by side**. It is difficult to visualize hyperblocks that do not overlap in n-D space also not overlapping in 2-D space of parallel coordinates. Figures 6 and 7 depict HBs that do not overlap in 9-D space but do so in parallel coordinates in 2-D, illustrating this challenge. We placed such HBs in separate Figs. 6 and 7, or side by side, to avoid drawing one HB on top of the other. The implementation of this side-by-side option in the VisCanvas 2.0 system is shown in Fig. 12.

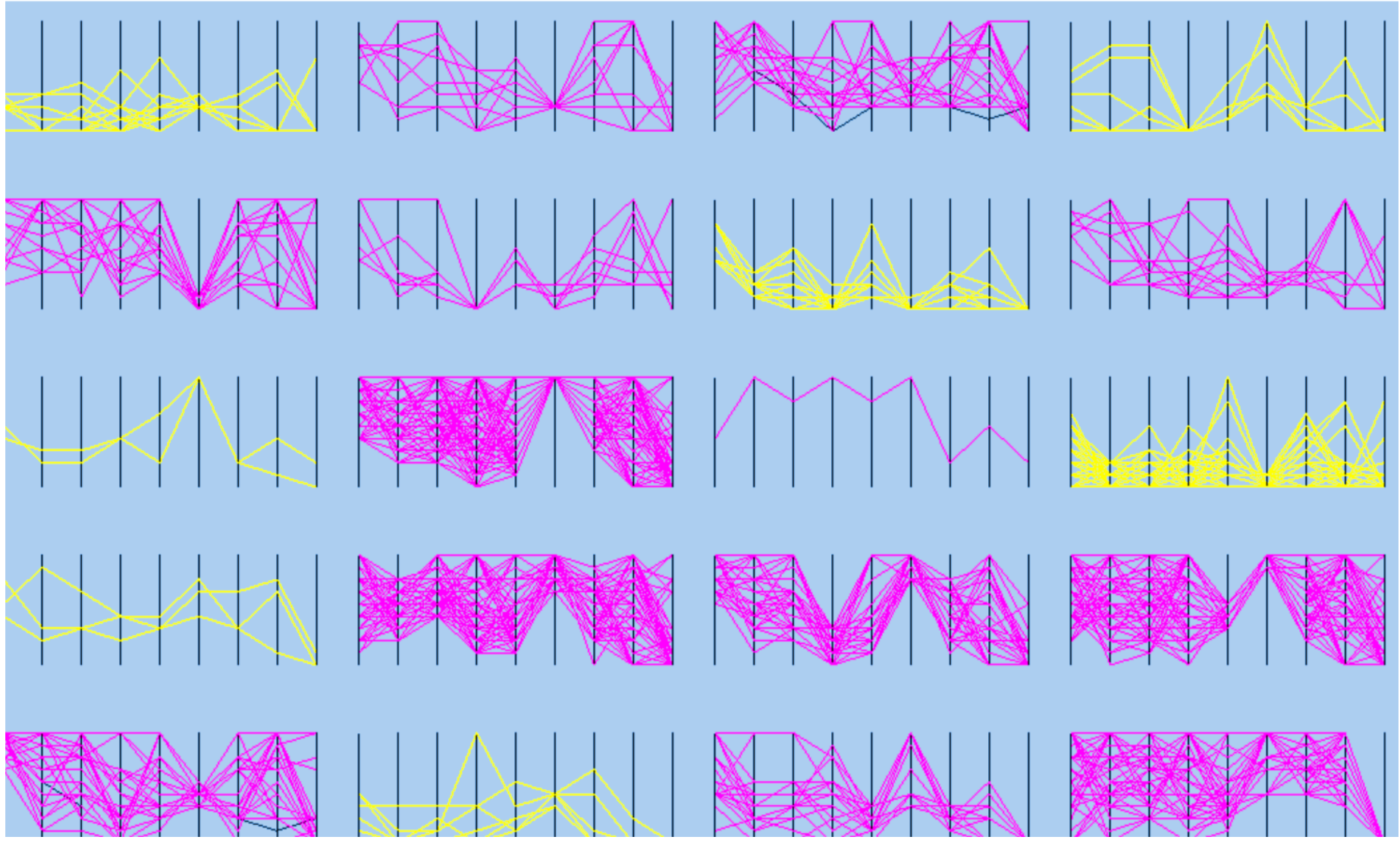

Fig. 12. All WBC discovered pure HBs of two classes side by side.

Another strategy is to display the data hyperblocks in **three dimensions**. First, the camera is set up to only to view two dimensions. The user can then tilt the camera to display the three-dimensional space, where each hyperblock is discernible in a different plane in the z-coordinate. HBs would be situated on the same plane if they overlap.

**Areas without overlap**. The following strategy involves displaying the variations in HB distribution in parallel coordinates in n-D space. First, all distinct hyperblock pairings are created. Then, these pairings are iterated over, and a rectangle is drawn on the dimensions where a given pair **do not overlap**.

This was implemented in several different ways as Fig. 13 shows. The more pairs that do not overlap in a coordinate, the darker the rectangle is. It is clear from this visualization in Fig. 13 that HBs less overlap in X6 coordinate.

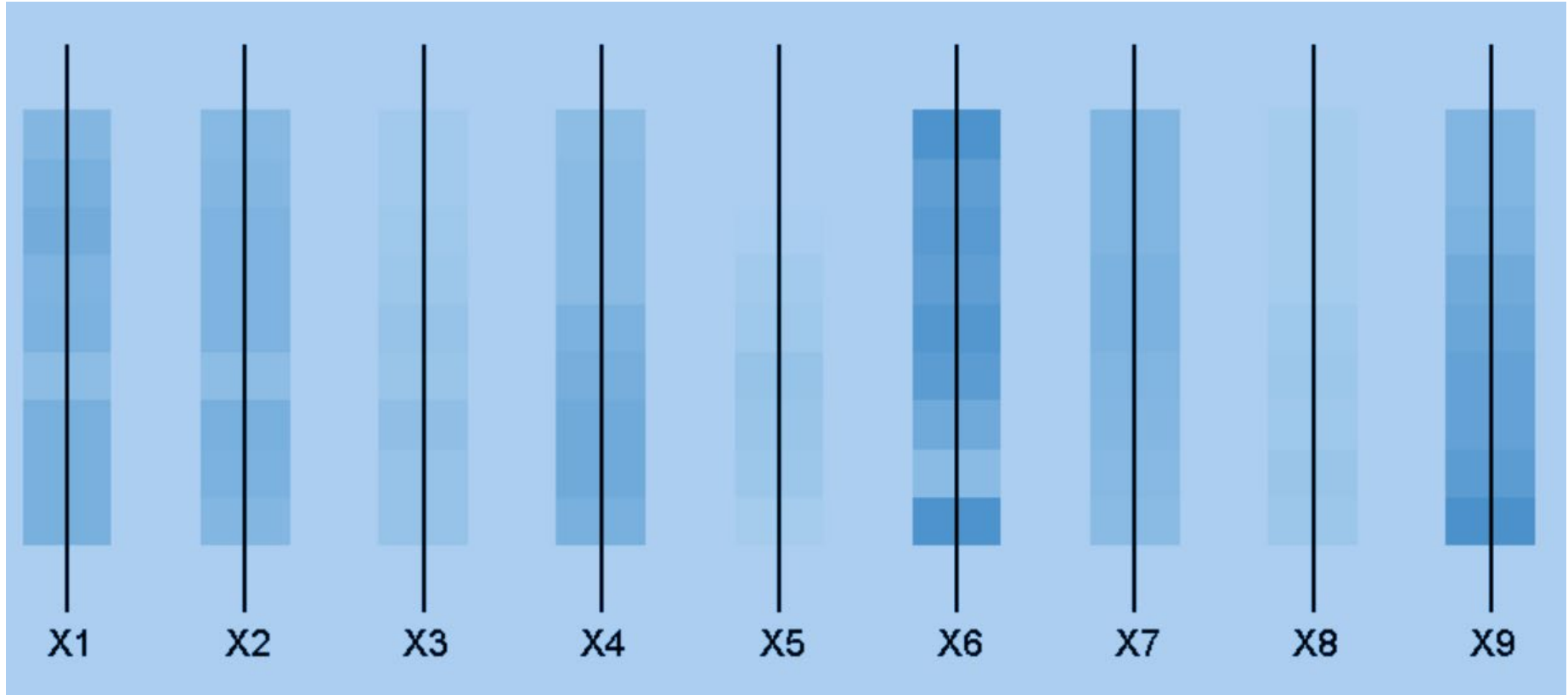

Figure 13. Locations where pairs of hyperblocks do not overlap.

### *2.5. Discovering and visualizing pairs of non-overlapping hyperblocks*

When polylines of opposing classes widely overlap and occlude one another, it becomes much more difficult to visually detect patterns of various classes in parallel coordinates. To improve the distinction computational and interactive techniques are required. One of the approaches is finding subsets of coordinates where the difference is quite clear and visualizing only data in those coordinates.

This search can be successful only if such subsets exist. The challenge is in multiplicity of HBs that represent each class. An individual pair of HBs from opposing classes can be quite distinct in some coordinates. However, as the number of HBs that represent classes increases, the likelihood of a subset of coordinates where all HBs not overlap goes down. Therefore, we search only for distinctive pairs of HBs and visualize each such pairs individually.

The successful search in WBC data that VisCanvas 2.0 executed is seen in the results below. To get to this point, VisCanvas takes the following automatic and interactive steps:

1. Auto-generate hyperblocks for opposing classes,
2. Chose hyperblocks which contain the most of data n-D points for each class,
3. Auto-reduce dimensions of these hyperblocks to dimensions where there these HBs do not overlap,
4. Visualize these HBs in reduced dimensions,
5. Ascertain a classification rule based on these HBs and evaluate its accuracy.

These steps resulted in 93.7% accuracy for these WBC data. Fig. 14 shows two such discovered HBs.

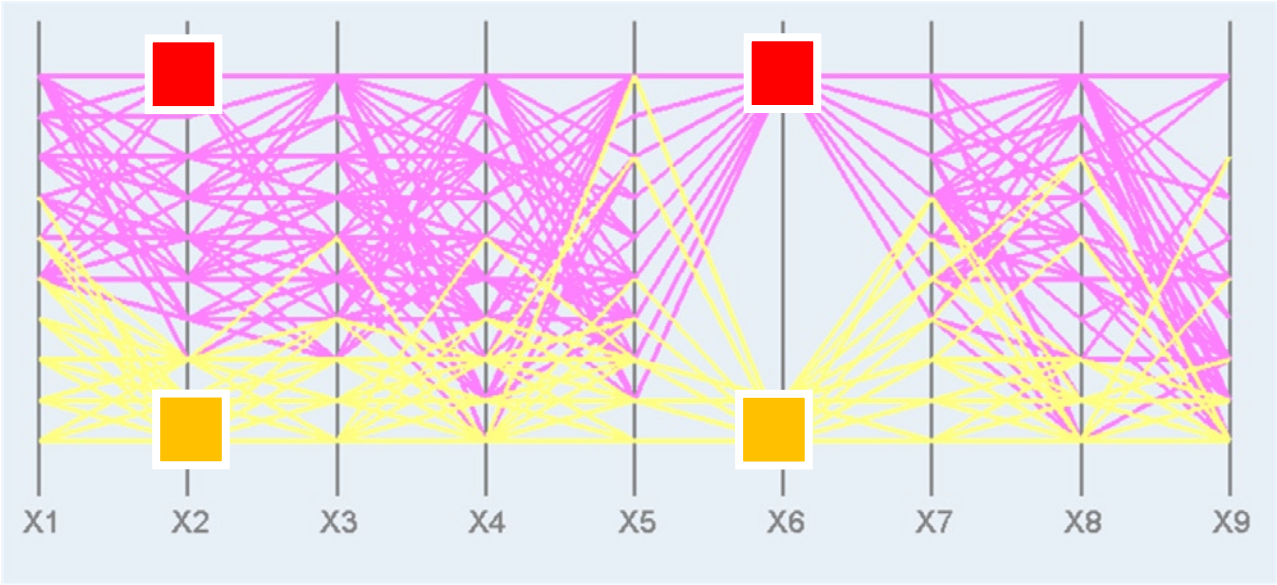

(a) Two non-overlapping hyperblocks (pink and yellow) with red and yellow squares marking separating coordinates.

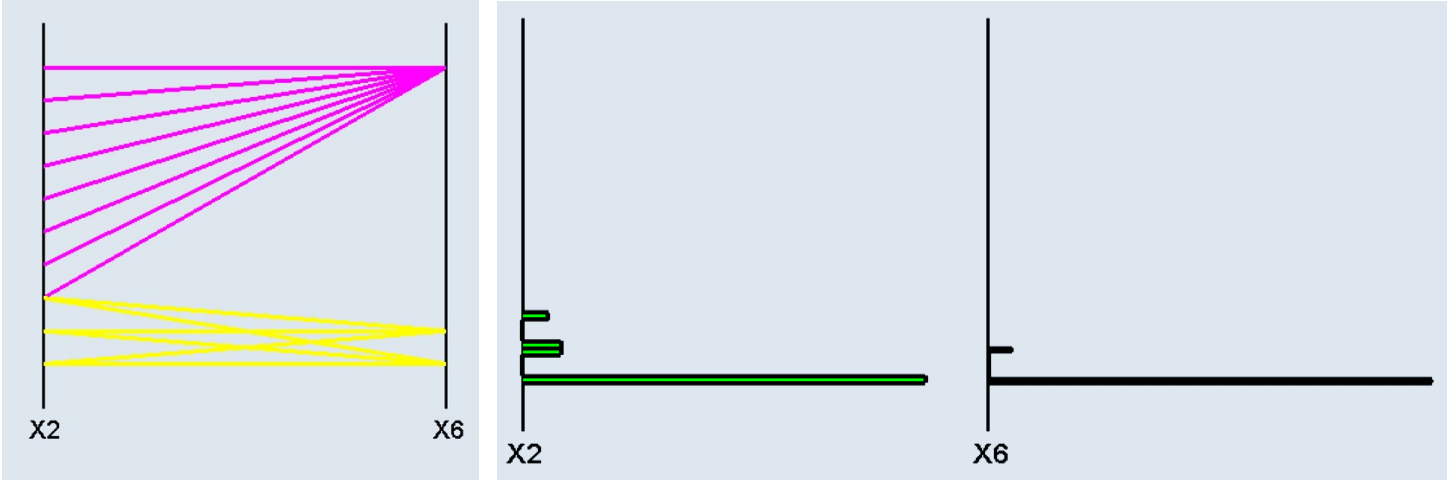

(b) Frequency visualization of yellow HB in most informative coordinates X2 and X6.

Fig. 14. Two hyperblocks of two classes in parallel coordinates.

## *2.6. Exploration and combination of hypelocks*

**Exploration of hyperblocks**. With VisCanvas 2.0, exploring hyperblocks begins by choosing a line that corresponds to an n-D point of interest. Then we build HB with this line as its center line and at a certain distance from it, such as 0.2. Hyperblocks with and without their lines are depicted in Figs. 15 and 16.

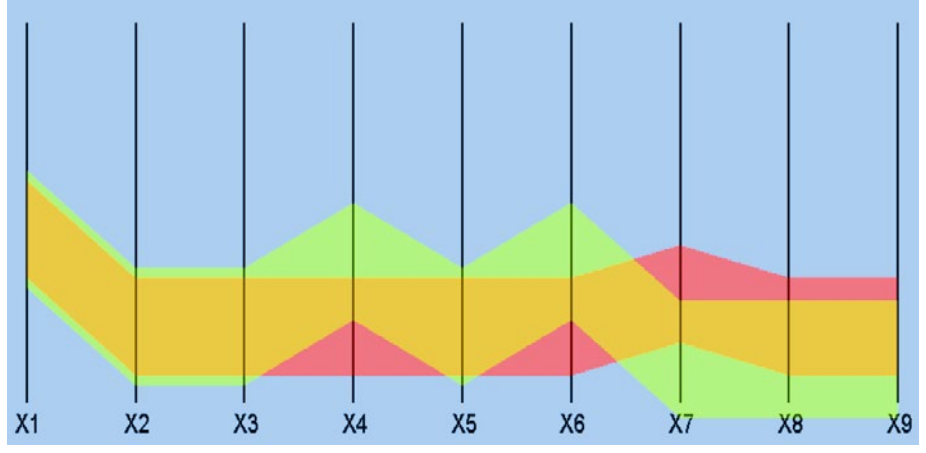

Fig. 15. Hyperblocks without lines.

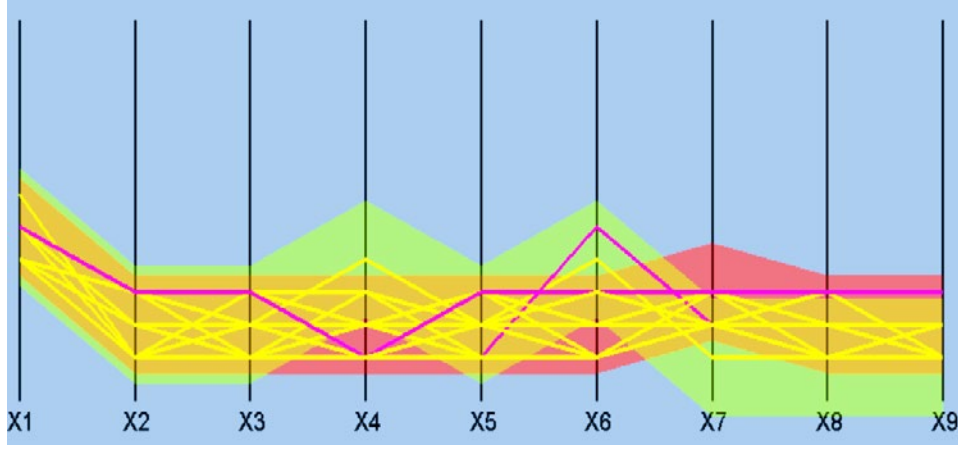

Fig. 16. Hyperbloicks with lines.

**Combination Modes**. We investigated three alternative ways to put hyperblocks together. The first mode (M1) examines the overlap of these HBs, or if all dimensions of each center are within range of one another. The second mode (M2) determines whether both HBs contain at least one point. Further verification of the presence of cases from other classes that may be in the joint hyperblock is necessary for both approaches to succeed. The third mode (M3) determines if one of each center's n-1 dimensions is inside one of the other's ranges and whether both HB n-1 dimensions are identical. This situation arise rarely. It is shown in Fig. 16.

In addition, setting a threshold allows showing hyperblocks that have a certain number of elements within them. Some hyperblocks **overlap**. Finding and removing overlaps and defining **joined** new HBs will minimize the number of them. HBs that have a **low purity** ratio are not good candidates to combine with other hyperblocks.

Therefore, it is better to hold them in their own. These blocks can then be drawn in a different color, toggled on/off, etc. Below we present algorithms and results of experiments for combining HBs and VisCanvas 2.0 features that support it.

VisCanvas 2.0 checks whether several n-D points produce the same HBs. It creates a single set of HBs from adjacent HBs. This is accomplished by comparing each prospective HB's dimension to each other HB's dimension.

They are considered as adjacent if the sum of all differences divided by 2 is less than or equal to the threshold value. Finally, VisCanvas 2.0 counts and displays the total number of unique HBs and the number of them with cases of multiple classes. It can be used to check for overfitting.

A user can view the ratio of HBs that have cases of multiple classes compared to the total number of unique HBs. For instance, it can show a HB with 93 cases from class 1, and 2 cases from class 2 in different colors. Multiple algorithms can be designed to combine HBs. One of them is presented below.

### *2.7. Linguistic description of visual patterns*

**Linguistic descriptions of hyperblocks combined with distribution visualization.** A linguistic pattern description is intended to make it easier for users to recognize the visual patterns. Frequently it is shorter and more "natural" for people and represents the entire n-dimensional space, not a sub-space that a specific HB resides in. With a straightforward UI, a user can activate a linguistic description.

While more precise divisions are being developed, these descriptions now point out data concentration in the lower, middle, and upper thirds of the coordinates. The current implementation allows merging descriptions of dimensions that are concentrated in the same thirds of them into a compact single description as Fig. 17 shows.

The distribution of n-D points might become distorted due to occlusion, as shown in Fig. 17b. A linguistic description prevents this contamination because it is based on a study of statistical data. It demonstrates the benefits of integrating visual representation, frequency, and language description. The n-D points, their frequency distribution, and a linguistic description of the hyperblock are all displayed in Fig. 17a.

Although the linguistic description in Fig. 17 is well organized in relation to the bottom, middle, and upper areas of data concentration in the respective coordinates, it is not a conventional natural language (NL) phrase. Modified linguistic descriptions are depicted in Fig. 18 as conventional NL phrases. Also, it is more succinct and enables the user to understand the language distinctions between the various classes.

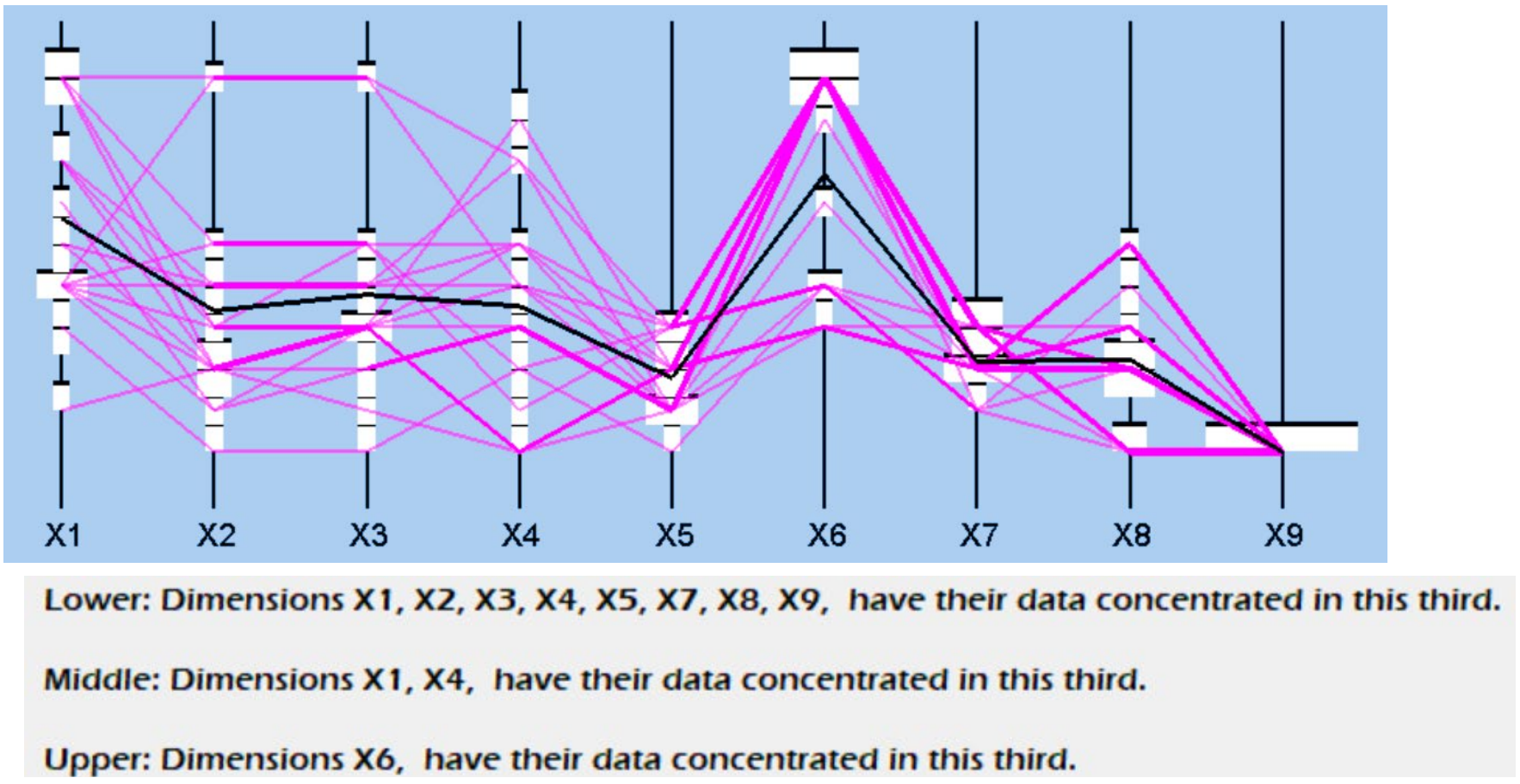

(a) A hyperblock with frequency pattern with its linguistic description.

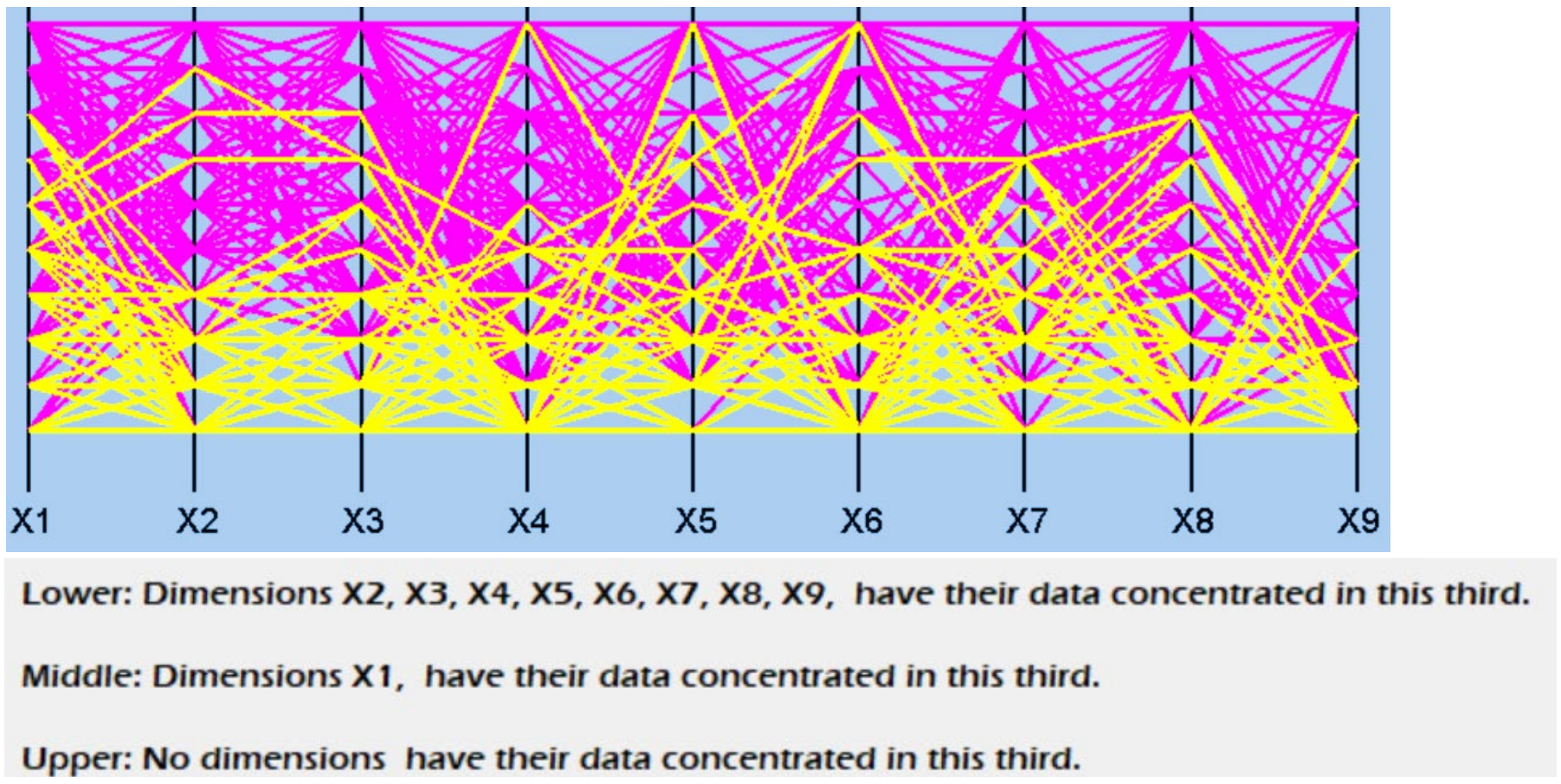

(b) All WBC data of both classes with linguistic description.

Fig. 17. WBC data visualized with linguistically annotated patterns.

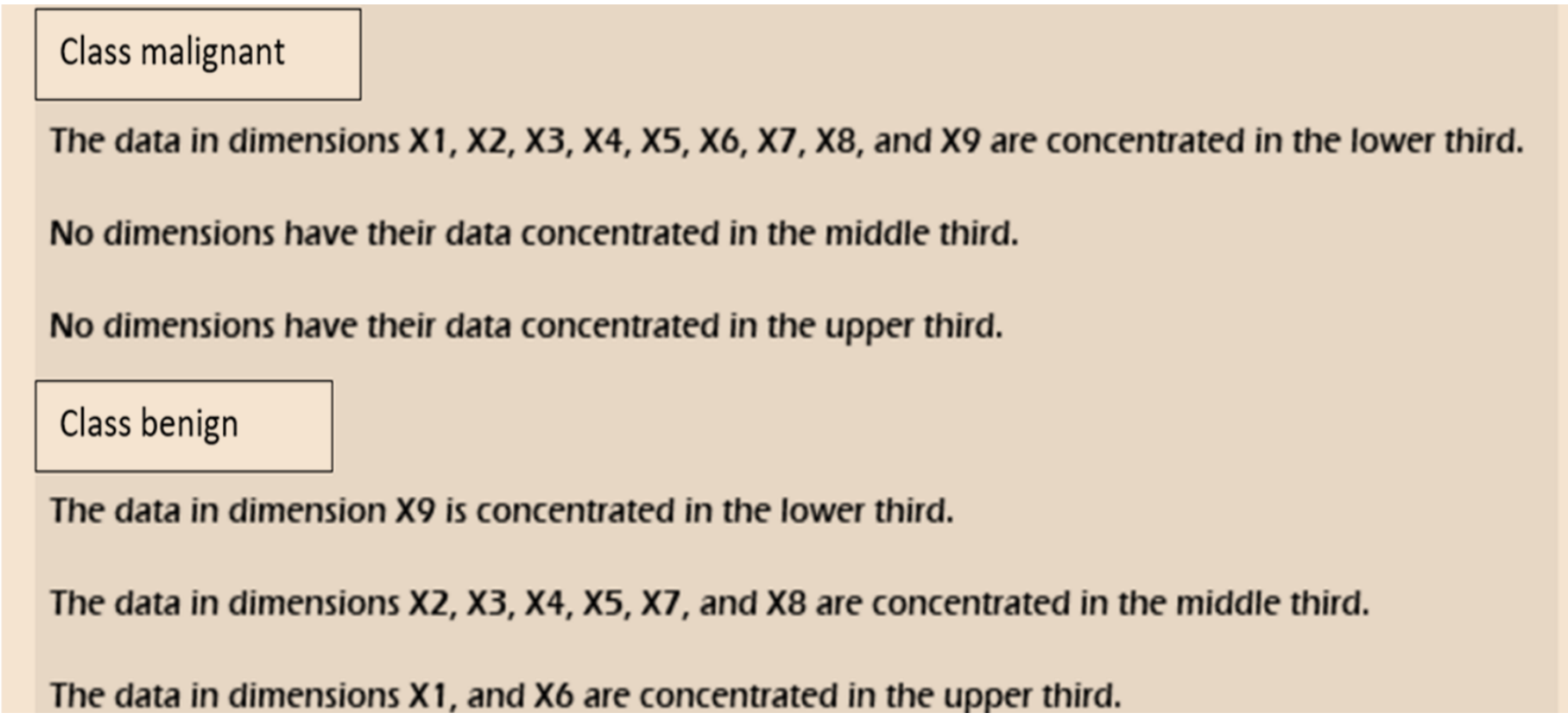

Fig. 18. Linguistic description of all WBC data.

## 3. SUPERVISED LEARNING IN PARALLEL COORDINATES

### *3.1 Hyperblock-based classification algorithm (HYPER)*

#### *3.1.1. Mixed and pure hyperblocks*

A supervised learning system called a **hyperblock-based classification (Hyper) algorithm** learns and displays interpretable rules in parallel coordinates. VisCanvas 2.0 software uses the Hyper algorithm. The core concept behind the Hyper algorithm is the classification of n-D points using n-D hyperblocks. First, it looks for dominant hyperblocks, which are regions where the majority of n-D points fall under one class. Let's say that HB1 dominates in class 1, HB2 dominates in class 2, and so on.

If a new n-D point **x** *belongs* to a respective HB where class $C_i$ dominates then **x** is classified to class $C_i$. If **x** does not belong to any dominant HB, then **x i**s classified to the *nearest HB*. If there are several competing HBs nearby, then the those nearest *k* HBs vote and **x** is classified to the class with *majority votes*. If there is not enough HBs to vote in the vicinity, the Hyper algorithm refuses to classify this n-D point **x**.

The uniqueness and advantages of the Hyper method are in its **integration** with parallel coordinate visualization and corresponding visual knowledge discovery. Its basic algorithmic ideas can be traced back to k-NN first developed in 1951 and algorithm of computation of estimates [14]. There are also recent works on hyperblocs relevant to k-NN [8, 9].

Below we introduce the notation and describe the Hyper algorithm in more details. Let $HB_i$ be a hyperblock where n-D points of *class $C_i$ dominate*. Also let $\{HB\}_k$ be a set of *k nearest HBs* for n-D point **x**, which *vote for class* $C_i$, i.e., *most* of these hyperblocks are class $C_i$ dominant HBs.

The Hyper algorithm learns rules in the following forms:

Rule 1: If n-D point **x** belongs to any dominant hyperblock $HB_i$, for class $C_i$, then **x** belongs to class $C_i$.

Rule 2: If $HB_i$ is a nearest hyperblock for n-D point **x,** then **x** belongs to class $C_i$.

Rule 3: If $\{HB\}_k$ is a set of $k$ nearest hyperblocks for n-D point **x** and the majority of $\{HB\}_k$ vote for class $C_i$, then **x** belongs to class $C_i$.

More formally these rules are:

$$R_1:\ \exists\, HB_i\, (\mathbf{x} \in HB_i) \Rightarrow \mathbf{x} \in C_i. \quad (3)$$

$$R_2: (HB_i \text{ is a nearest HB for } \mathbf{x}) \Rightarrow \mathbf{x} \in C_i \quad (4)$$

$$R_3: (\{HB\}_k \text{ is } k\text{-}NN \text{ HBs for } \mathbf{x})\&(\text{Vote}(\{HB\}_k)=C_i) \Rightarrow \mathbf{x} \in C_i \quad (5)$$

The **major steps of Hyper algorithm** to learn $R_1$-$R_3$ are:
Step H1: Split data to *training data* $T_r$ and *validation data* $V_d$ (e.g., by 10-fold cross validation) and split $Tr$ to $T_{rh}$ for learning HBs and $T_{rk}$ learning the number of nearest neighbors HBs.
Step H2. Learn *dominant HBs*.
Step H3. Learn the number $k$ of nearest neighbors HBs.
Step H4. Generate rules $R_1$-$R_3$.
Step H5. Validate rules $R_1$-$R_3$ using validation data $V_d$.

Below we present Steps H2 and H3 in more details.
Step H2. Learn *dominant HBs*.

Step H2.1. Construct all pure HB on training data $T_r$ by the algorithm presented in Section II.B.

Step H2.2. Collect all n-D points $\{\mathbf{b}_j\}$ that belong to *single-point HBs* (HBs without other n-D points).

Step H2.3. For each $\mathbf{b}_j$ find a nearest HB. If the class of $\mathbf{b}_j$ and the nearest pure HB is the same, record it as a positive classification, else record it as negative classification.

Step H2.4. Attempt to combine HBs to larger dominant HBs by adding small HBs of other classes to adjacent large HBs keeping HB purity above in a given threshold $T$.

Step H3. Learn the number of nearest neighbors HBs $k$.

Step H3.1. Set up an interval of possible number $k$ of nearest HBs.

Step H3.2. Loop through $k$ is this interval to find $k$ with most accurate voting classification for all n-D points in $T_{rk}$. If predefined accuracy threshold Q is reached, finish learning process, else leave n-D points in $T_{rk}$ unclassified and exit.

### *3.1.2. Sets of hyperblocks as generalized decision trees*

Below we discuss learning **simple dominant HBs** such as common in **decision trees**. According to the HB definition it satisfies inequalities $\| x_i\text{-}c_i \| \leq L_i/2$ for a respective *center* n-D point $\mathbf{c}=(c_1,c_2,\ldots,c_n)$ and *lengths* $\boldsymbol{L}=(L_1, L_2,\ldots, L_n)$.

Consider an example of a hyperblock $\|x_i\text{-}c_i\| \leq c_i$ in parallel coordinates that are in [0,10] interval each. This inequality is true for all non-negative $x_i \leq 2c_i$, in other words, $x_i \in [0,2c_i]$. Such simple HB needs only $c_i$ values that identify its center. Similarly, consider a complimentary HB where $x_i \in (2c_i,10]$, i.e., $x_i > 2c_i$. Its center $\boldsymbol{h} = (h_1,h_2,\ldots,h_n)$ is defined by $h_i=(10\text{-}2c_i)/2$. As we see these HBs are defined by a simple set of inequalities.

The Hyper algorithm allows discovering and visualizing hyperblocks of the type of branches of decision trees (DTs). It is based on a direct link between DTs and hyperblocks (HBs). Consider, a branch of DT: $x_1>5$ & $x_2< 6$ & $x_3> 2$ that is labeled, say, by class 1. Assume that all coordinates $X_i$ are in [0.10] interval, then HB for this branch is defined by three intervals:

$$x_1 \in (5,10],\ x_2 \in [0,6) \text{ and } x_3 \in (2,10].$$

Alternatively, instead of starting from a DT, we can start from a hyperblock. Let HB be given by three intervals $x_1 \in (5,7)$, $x_2 \in (3,6)$, and $x_3 \in (2,4)$ within interval [0,10]. It is equivalent to the DT branch:

$$x_1>5 \ \&\ x_1<7 \ \&\ x_2< 6 \ \&\ x_2> 3 \ \&\ x_3> 2 \ \&\ x_3< 4 \quad (6)$$

These examples show a 1:1 mapping between a DT branch and a hyperblock without loss of generality.

A user can create a DT outside of VisCanvas 2.0 and then convert each DT branch to a respective HB to be visualized in VisCanvas 2.0. Alternatively, a set of HBs produced in VisCanvas 2.0 can be viewed as a set of DTs as follows. Assume that each discovered HB is represented by a DT branch like (4). Can we combine these branches to a single tree? It is possible only in a special situation when branches have a common root, like in the following example. Say, branch *A* contains inequality $x_i \leq T$, and branch *B* contains the opposite inequality $x_i>T$, then coordinate $X_i$ can serve as a common root for *A* and *B*. Thus, outside of special cases, a set of HBs (HB "forest") is a more general model than a DT model. A set of HBs removes a limitation of a single DT requiring a root.

## 3.2. *WBC case study*

### 3.2.1. *Learning of hyperblocks with all data*

WBC data have been used to compare the Hyper method to the ID3 decision tree algorithm used in VisCanvas 2 and Scikit-learns, respectively. First, we try to discover a hyperblock $HB_i$ that is dominant for class $C_i$ and then to build a rule on all WBC data:

$$\text{If } \mathbf{x} \in HB_i \Rightarrow \mathbf{x} \in \text{class } C_i \text{ dominant in } HB_i \quad (7)$$

Investigating the rule's accuracy is part of the study. The best-case accuracy estimate on this WBC data with HBs can be achieved by using all WBC data for training (discovering HBs) without reserving a subset for validation.

Table 1 shows the HBs generated by the Hyper algorithm when using a 0.2 distance from the center of HB in WBC data normalized to [0,1]. Total 22 HBs have been produced: 20 pure and 2 mixed. Out of 20 pure HBs 7 belong to the class B (benign) and 13 to class M (malignant). Two mixed HBs are very different. One is almost pure class M dominant, 92/1, but the second HB is not with 2 cases and ratio 1/1. In this situation when $HB_i$ is not dominant for any class, all n-D points of this HB are classified to class M (malignant) to avoid more dangerous misclassification to benign class.

The number of n-D points in each HB also varies very significantly from 404 n-D points to a single n-D point for class B, and from 76 n-D points to single n-D point for class M (see Table 2). The produced HBs overlap due to the version of the algorithm used. A more elaborated version allows removing overlaps. Having only two mixed HBs with ratios 92/1 and 1/1 the Hyper algorithm misclassified only two benign cases with total accuracy 681/683, i.e., 99.70%.

Table 1. Hyperblocks with using a 0.2 center distance with the WBC dataset.

| Type of HCs | Number of HCs | Number of n-D points in HCs |
|---|---|---|
| Pure | 20 | 588 |
| Mixed | 2 | 95 |
| Total | 22 | 683 |

Table 2. Number of n-D points in pure overlapped HBs for WBC dataset.

| class | Number of n-D points in hyperblock | | | | | | | | | | | |
|---|---|---|---|---|---|---|---|---|---|---|---|---|
| B | 404 | 392 | 92 | 34 | 16 | 5 | 4 | 1 | | | | |
| M | 76 | 44 | 42 | 39 | 34 | 25 | 18 | 18 | 12 | 9 | 7 | 1 |

Note, this rule relies on several very small HBs, e.g., 7 HBs include 9 or less n-D points, which can be interpreted as overfitting. To avoid it we can refuse to classify n-D points that are in these HBs, i.e., remove these HBs from the rule. These HBs contain 29 cases. It will decrease the number of cases covered by the rule to 683-29=654 (recall 95,75%) with precision 99.85% (653/654) that is slightly greater than the accuracy 99.7%. The increase the threshold to 25 cases in a HB will remove 11 HBs with total 91 n-D points, resulted in recall of 86.68% and precision of 99.83. These precisions set up two other best-case benchmarks for HB-based algorithms on WBC data.

#### *3.2.2. Supervised learning with training and validation data*

The WBC dataset was subjected to **10-fold cross validation** using VisCanvas 2.0 and Scikit-learns ID3 decision tree implementation. For each 10-fold split, the Hyper algorithm in VisCanvas 2.0 automatically built hyperblocks, and rules (1) through (3) derived from these HBs were evaluated on the validation folds. Three versions of *k*-NN were explored for each n-D point **x** with *k*=1,3, 5:

(N1) the distance from **x** to the *center n-D points* of discovered hyperblocks,
(N2) the distance from **x** to the *mean n-D point* of discovered hyperblocks,
(N3) the distance from **x** to the *nearest n-D point*

For N1-N3 with pure HBs, Table 3 presents the results that are similar to each other with average accuracy above 95% and reaching 97.61%. The version N3 produced the lowest results. It means that this method is more sensitive to the distribution of closest points, while methods N1 and N3 based on means and centers are more robust.

The average accuracy for the ID3 decision tree is 92.85%, with min equal to 89.86% and max equal to 94.58%. All of them are **below** the averages for N1-N3 of the *k*-NN hyperblock algorithm Hyper, which shows the advantages of hyperblocks for this dataset. The decision tree has size depth 8, it contains 23 branches which are larger than the average number of HBs which is 18.1 in the worst case.

Likewise, for N1-N3 with mixed HBs, Table 3 presents the results with similar accuracy scores. This could be because the number of HBs has been reduced to approximately 6 in comparison with the pure HBs average of 18. These HBs cover more space and it is easier to place n-D points within them for classification. As a result, the accuracy would be consistent and there would be no requirement for N1, N2, or N3 in the k-NN version of the algorithms. Moreover, it consistently outperformed decision tree averages, with the exception of minimal accuracies, where it lagged behind the decision tree.

Table 3. Summary of experiments with WBC data using pure and mixed HBs.

| Model type | 10-fold accuracy | | | average # of HB |
|---|---|---|---|---|
| | average | min | max | |
| *k*=1 | | | | |
| N1 pure | **95.52** | 91.04 | 100 | 16.9 |
| N2 pure | 95.22 | 89.55 | 100 | 16.9 |
| N3 pure | 93.58 | 86.57 | 100 | 16.9 |
| N1 mix | 93.43 | 86.57 | 97.01 | 5.4 |
| N2 mix | **93.58** | 86.57 | 97.01 | 5.4 |
| N3 mix | 93.28 | 86.57 | 98.51 | 5.4 |
| *k*=3 | | | | |
| N1 pure | 96.57 | 92.54 | 98.51 | 17.3 |
| N2 pure | **97.61** | 94.03 | 100 | 17.3 |
| N3 pure | 94.78 | 91.04 | 97.01 | 17.3 |
| N1 mix | **93.88** | 88.06 | 98.51 | 7.1 |
| N2 mix | **93.88** | 88.06 | 98.51 | 7.1 |
| N3 mix | 93.73 | 88.06 | 97.01 | 7.1 |
| *k*=5 | | | | |
| N1 pure | **96.42** | 92.54 | 100 | 18.1 |
| N2 pure | **96.42** | 92.54 | 100 | 18.1 |
| N3 pure | 94.78 | 89.55 | 100 | 18.1 |
| N1 mix | **96.12** | 91.04 | 98.51 | 6.4 |
| N2 mix | **96.12** | 91.04 | 98.51 | 6.4 |
| N3 mix | 95.82 | 91.04 | 98.51 | 6.4 |

For pure and mixed hyperblocks applied in the Hyper method, Table 4 displays confusion matrices for the fold that comes the closest to the average accuracy in 10-fold cross validation for *k*=1,3,5.

Table 4. Confusion matrixes for the fold that is closest to the average accuracy.

| | N1 | B | M | N2 | B | M | N3 | B | M |
|---|---|---|---|---|---|---|---|---|---|
| **Pure hyperblocks** | | | | | | | | | |
| *k*=1 | N1 | B | M | N2 | B | M | N3 | B | M |
| | B | 44 | 0 | B | 44 | 0 | B | 44 | 0 |
| | M | 3 | 20 | M | 3 | 20 | M | 4 | 19 |
| *k*=3 | N1 | B | M | N2 | B | M | N3 | B | M |
| | B | 43 | 1 | B | 43 | 1 | B | 44 | 0 |
| | M | 1 | 22 | M | 1 | 22 | M | 3 | 20 |
| *k*=5 | N1 | B | M | N2 | B | M | N3 | B | M |
| | B | 42 | 2 | B | 43 | 1 | B | 44 | 0 |
| | M | 0 | 23 | M | 2 | 21 | M | 3 | 20 |
| **Mixed dominant hyperblocks** | | | | | | | | | |
| *k*=1 | N1 | B | M | N2 | B | M | N3 | B | M |
| | B | 44 | 0 | B | 42 | 2 | B | 42 | 2 |
| | M | 3 | 20 | M | 2 | 21 | M | 3 | 20 |
| *k*=3 | N1 | B | M | N2 | B | M | N3 | B | M |
| | B | 43 | 1 | B | 43 | 1 | B | 44 | 0 |
| | M | 1 | 22 | M | 1 | 22 | M | 3 | 20 |
| *k*=5 | N1 | B | M | N2 | B | M | N3 | B | M |
| | B | 41 | 3 | B | 43 | 1 | B | 43 | 1 |
| | M | 0 | 23 | M | 2 | 21 | M | 3 | 21 |

Table 5. Number of n-D points in nodes of ID3 for WBC dataset.

| Cases | Number of n-D points of two classes in the node | | | | | | | | | | | | |
|---|---|---|---|---|---|---|---|---|---|---|---|---|---|
| *Node* | *1* | *2* | *3* | *4* | *5* | *6* | *7* | *8* | *9* | *10* | *11* | *12* | *13* |
| Class B | 397 | 364 | 33 | 363 | 1 | 15 | 18 | 363 | 0 | 1 | 0 | 15 | 0 |
| Class M | 217 | 9 | 208 | 2 | 7 | 3 | 205 | 1 | 1 | 0 | 7 | 0 | 3 |
| *Node* | *14* | *15* | *16* | *17* | *18* | *19* | *20* | *21* | *22* | *23* | *24* | *25* | *26* |
| Class B | 15 | 3 | 362 | 1 | 9 | 6 | 2 | 1 | 1 | 0 | 8 | 1 | 5 |
| Class M | 149 | 156 | 0 | 1 | 4 | 45 | 9 | 147 | 0 | 1 | 0 | 4 | 16 |
| *Node* | *27* | *28* | *29* | *30* | *31* | *32* | *33* | *34* | *35* | *36* | *37* | *38* | *39* |
| Class B | 1 | 2 | 0 | 1 | 0 | 0 | 1 | 4 | 1 | 0 | 1 | 2 | 0 |
| Class M | 29 | 2 | 7 | 24 | 123 | 4 | 0 | 16 | 0 | 22 | 7 | 0 | 2 |
| *Node* | *40* | *41* | *42* | *43* | *44* | *45* | *46* | *47* | *48* | *49* | *50* | *51* | *52* |
| Class B | 0 | 1 | 3 | 1 | 1 | 0 | 1 | 0 | 3 | 0 | 1 | 0 | 0 |
| Class M | 20 | 4 | 4 | 12 | 1 | 6 | 0 | 4 | 1 | 3 | 1 | 11 | 1 |
| *Node* | *53* | *54* | *55* | *56* | *57* | | | | | | | | |
| Class B | 1 | 0 | 3 | 0 | 1 | | | | | | | | |
| Class M | 0 | 1 | 0 | 1 | 0 | | | | | | | | |

Comparable ID3 Decision tree on training data is shown in Table 5 and Fig. A1 in the appendix. The confusion matrix for these ID3 decision trees on the representative validation fold is shown in Table 6, where the number of errors for the cancer class is higher than that obtained by the Hyper approach.

Table 6. ID3 Confusion matrix.

| **ID3** | **2** | **4** |
|---|---|---|
| **B** | 45 | 2 |
| **M** | 5 | 17 |

The **complexity** of HBs vs. DT can be measured by comparing the number of values needed to store HBs and DT. Our estimates had shown that this DT will require significantly more values with ratio above 1.5.

### *3.2.3. Merger*

The steps of the Merger Hyperblocks **(MHyper**) algorithm (see section 2) were performed on the WBC dataset in conjunction with 10-fold cross validation. The average number of HBs created for each fold was 17.3. Four of the folds contain HBs that include only a single n-D point. Three folds had one HB with only a single n-D point and another fold had two HBs with only a single n-D point. To help this approach gain more generality it was adapted to also consider **dominant** HBs.

Next, we reduced the number of single points HBs including impure combinations. All steps of MHyper algorithm were performed with a threshold of 10% impurity. It produced 5 HBs in average for each fold without any single n-D point HB. This approach of defining a threshold for the percentage of n-D points from an opposite class allowed a balance between generalization and accuracy to be made. Table 7 shows the resulting 5 HBs in a representative fold. The merger algorithm allows overlap of merged HBs, which is seen in table 7. The average number of HBs decreases about 3 times from the initial number of pure HBs. These HBs have impurity less than 10% (see Table 7).

We also compared complexity of HBs vs. the decision tree by using the number of cases in the smallest merged HB. This merged HB contains 53 cases, which is 7.76% of the overall data. The DT's minimum number of n-D points within its nodes is 1. This is 0.15% of the overall data. It has 10 of these nodes with a single element within it. It also has 46 nodes that contain less than 53 cases. So, 80.7% of the DT's nodes are less generalized than the HB's least generalized block. The merger of HBs and pruning of DTs leads to both higher generalization and higher error rate that is controlled by the impurity threshold for HBs.

Table 7. Results of MHyper for impure HBs for a representative fold.

| | Number of n-D points of two classes in HBs | | | | |
|---|---|---|---|---|---|
| **Hyperblock** | **1** | **2** | **3** | **4** | **5** |
| Cases of Class B | 25 | 0 | 439 | 432 | 12 |
| Cases of Class M | 226 | 53 | 47 | 30 | 137 |
| impurity | 9.9% | 0 | 9% | 6.5% | 8% |

### *3.2.4. Simple hyper-boxes and dimension reduction*

Below we present a rule discovery by the Hyper algorithm with simple HBs of type of decision tree branches. The simplest discovered rule on WBC data is a **one-dimensional rule**

$$\text{if } x_6 < 3 \text{ then class 2 (benign) else class 4 (malignant)} \quad (8)$$

This rule has accuracy of 91.22% with correctly classified 623 cases out of 683 total cases.

A **two-dimensional rule** is

$$\text{if } x_6<3 \;\&\; x_8<4 \text{ then class B (benign) else class M} \tag{9}$$

This rule has accuracy of 93.85% with correctly classified 641 cases out of 683 total cases.

A **three-dimensional rule** is

$$\text{if } x_6<3 \;\&\; x_8<4 \;\&\; x_5 < 6 \text{ then class 2 (benign) else class M} \tag{10}$$

This rule has accuracy of 94.58% with correctly classified 646 cases out of 683 total cases. These rules have been discovered in a **dimension reduction process** in VisCanvas 2.0 where a user can toggle coordinates on and off in the "Dimension" tab of the settings to reduce data dimension. This is resulted in reduction of WBC dataset from 9-D down to the 3-D where patterns were perceived previously visually, the resulting patterns become even more obvious.

Two clearly separated hyperblocks of opposite classes are displayed in Fig. 14. It shows the pink hyperblock that contains 95 malignant cases, and the yellow hyperblock that contains 404 benign cases. It is a large majority of all data and the mirror like difference of these two HBs of two classes makes the pattern easier to distinguish. This is in line with rule (10) that involves the same 3 coordinates. The general form of the rules above is

$$\text{if } x_i<T_i \;\&\; x_j<T_j \;\ldots\; x_k<T_k \text{ then class B (benign) else class M.}$$

## 4. VISCANVAS 2.0 SOFTWARE FEATURES

### 4.1. Interaction features

**Observing and analyzing n-D data**, hypercubes, and hyperblocks in parallel coordinates is the major goal of VisCanvas 2.0. Fig. 19 shows two hyperblocks generated in VisCanvas 2.0 along with the frequency distribution of values.

The data table menu includes options for analyzing hyperblock. On opening a file, a hyperblock is created around every single data point for the use in this table. The table displays the class structure within the hyperblock. A user can select/deselect HBs for viewing. It allows a user to view the space around n-D points of interest. Fig. 20 shows an example of the data table and respective user interface.

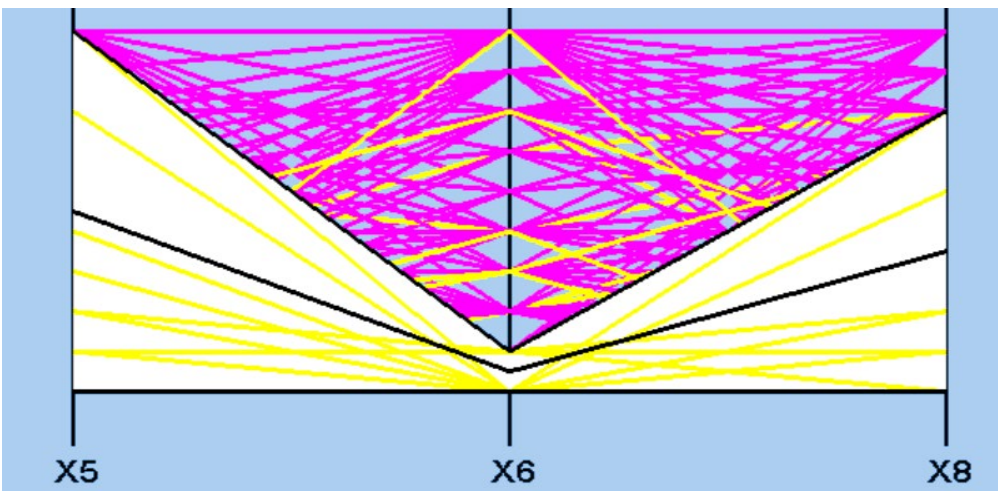

(a) HB of yellow class with all data on background

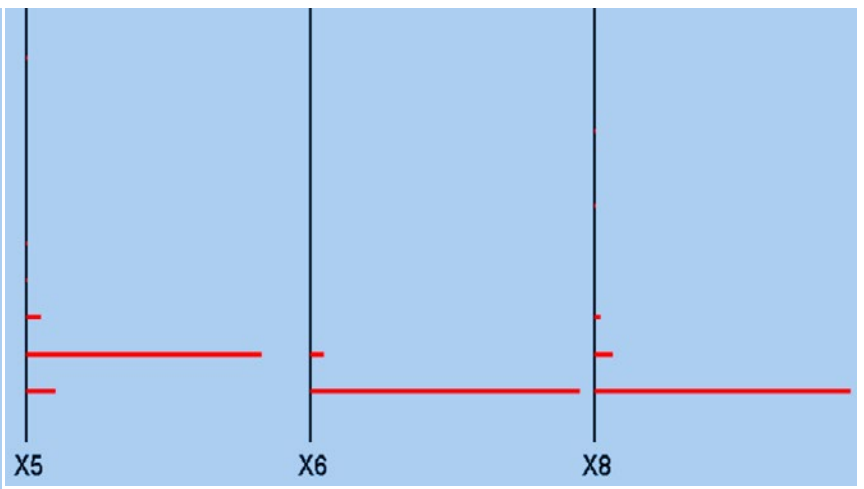

(b) Frequencies of yellow hyperblock

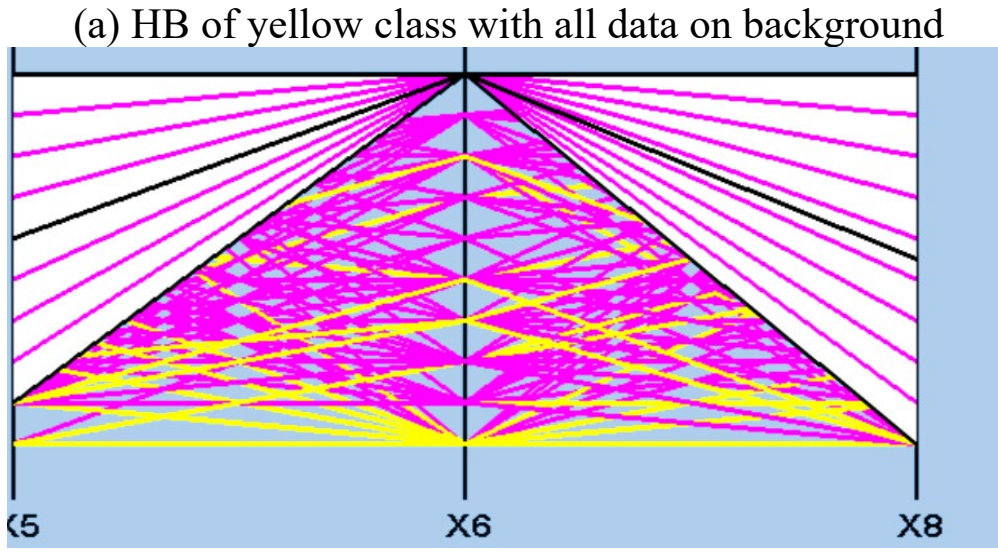

(c) HB of pink class with all data on background.

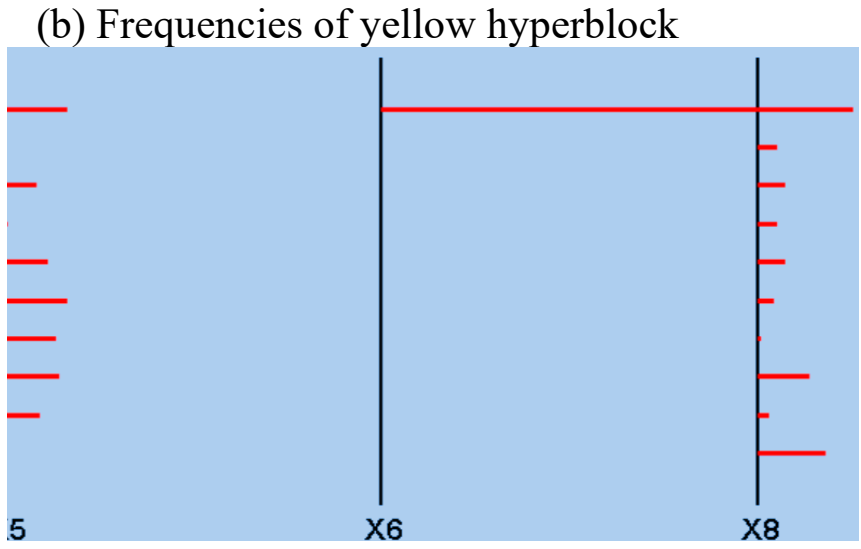

(d) Frequencies of p*i*nk hyperblock.

Fig. 19. Two distinct HBs of two classes with frequency distributions of n-D points in these HBs in respective coordinates.

**Creating subsets of n-D points**. VisCanvas 2.0 has a feature that allows creating subsets of n-D points. It includes a tool strip menu (see Fig. 20) and can be accessed through the subsets drop down. It has a column of checkboxes to make n-D points visible or not. Another column indicates the number of n-D points of other classes this n-D point crosses. The remaining columns show the values of coordinates of n-D points.

This table is populated by choosing one of the classes from the dropdown menu above it. The "All" button selects/de-selects all points in the given class. The "Apply" button sets the visibility according to checked checkboxes.

The **hyperblock selection utilities** have their own tab in the settings page and buttons with icons. The GUI options panel contains the data for the currently selected hyperblock and updates when the currently selected hyperblock changes. It also allows the user to go directly to a point or hyperblock by using its index.

The min, center, and max lines of a HB can also be toggled on and off. Fig.21 shows a selected hyperblock in main VisCanvas GUI window. The black line is its center **c**. the panel on right shows the number of n-D points on this hyperblock.

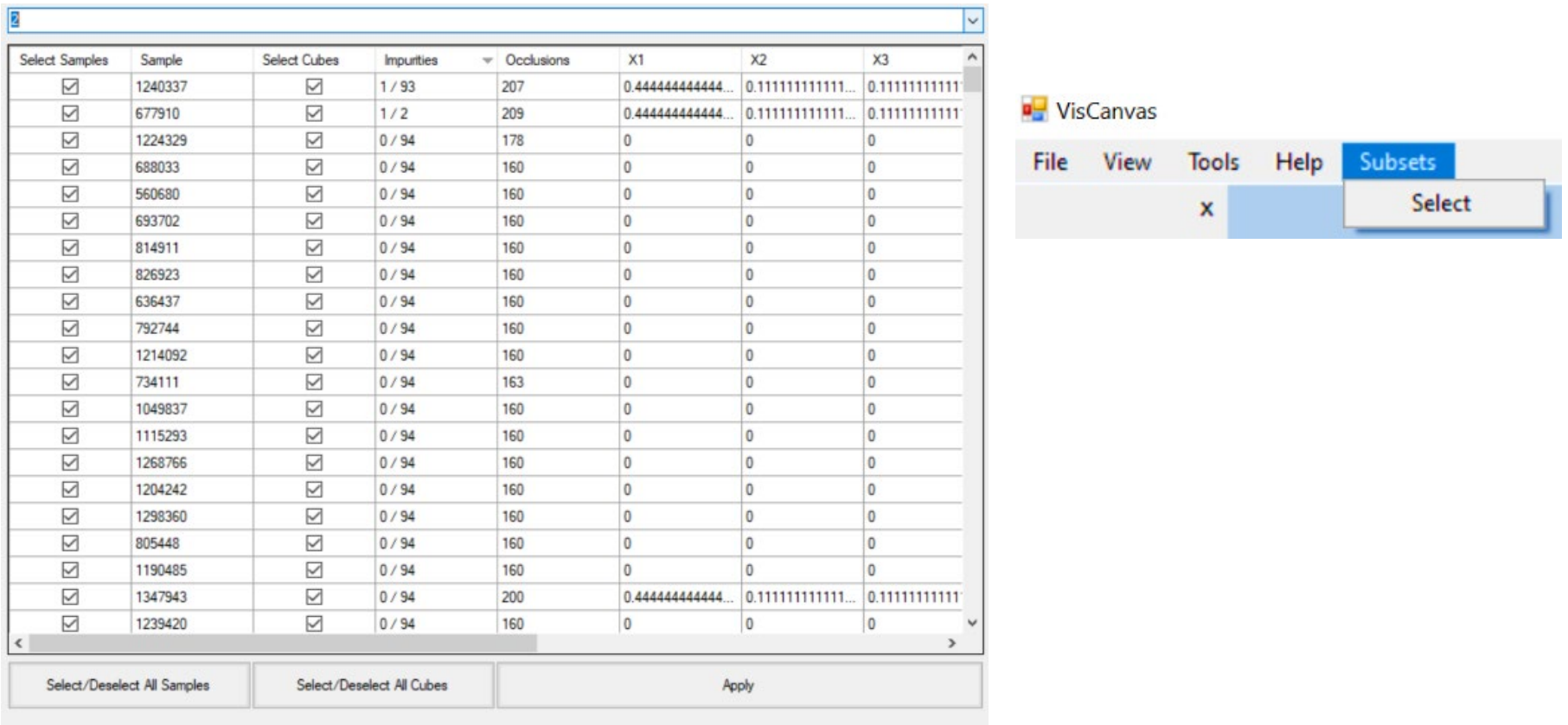

Fig.*20*. Accessing the subsets feature via the toolstrip menu dropdown and *data table* example

**Toggled drawings**. The user can independently toggle different lines on/off (see Fig. 22). This allows the user to control which visual data to observe instead of showing all or none of it.

The minimum/maximum lines are black and represent the lower and upper borders of the HB. The center line is also black and represents the median of the HB, its center. The selector line represents the currently selected line and is black by default, but the color can be changed by the user. See Fig. 21 with white background.

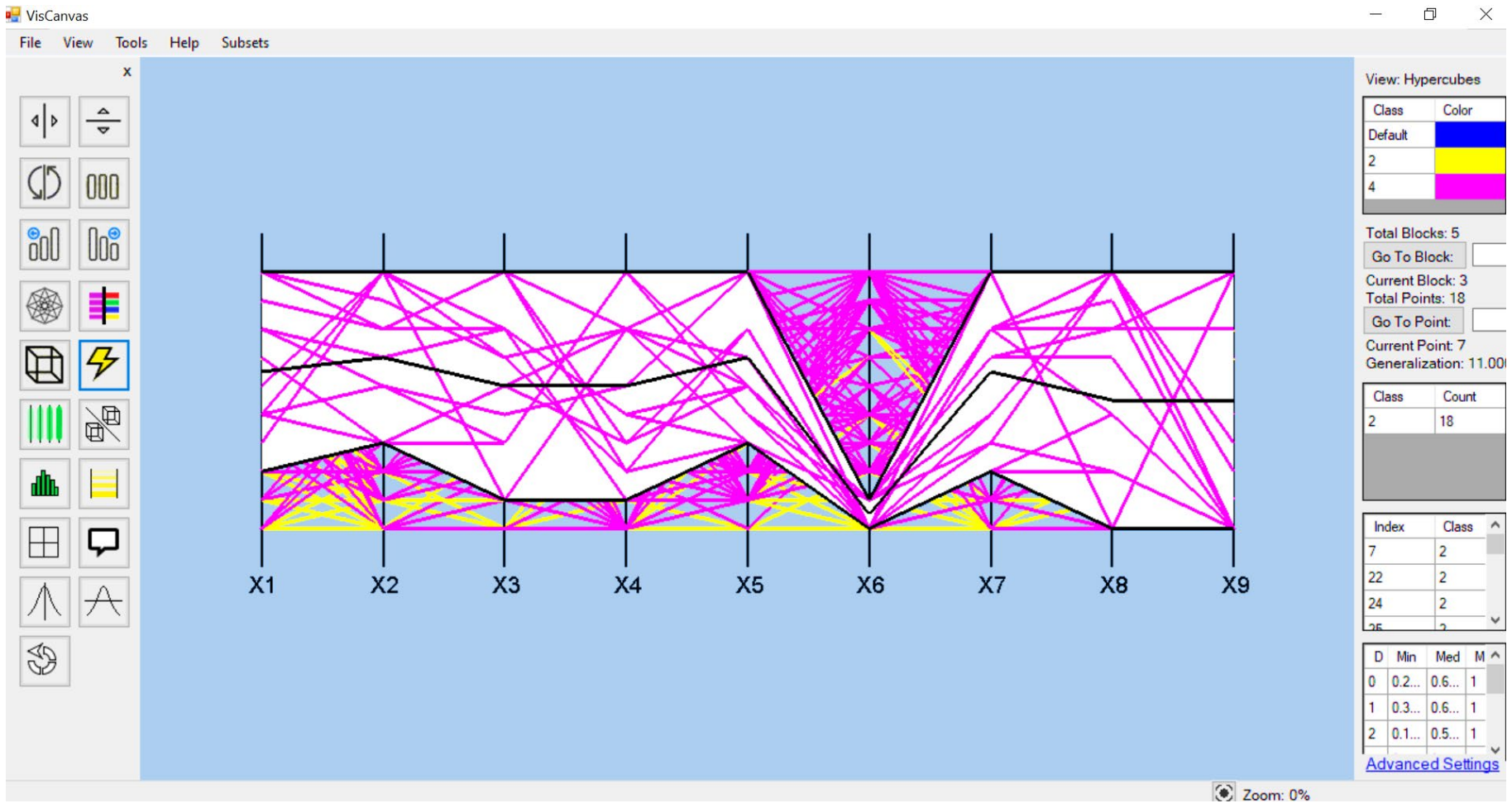

Fig.21. A hyperblock in main VisCanvas GUI window.

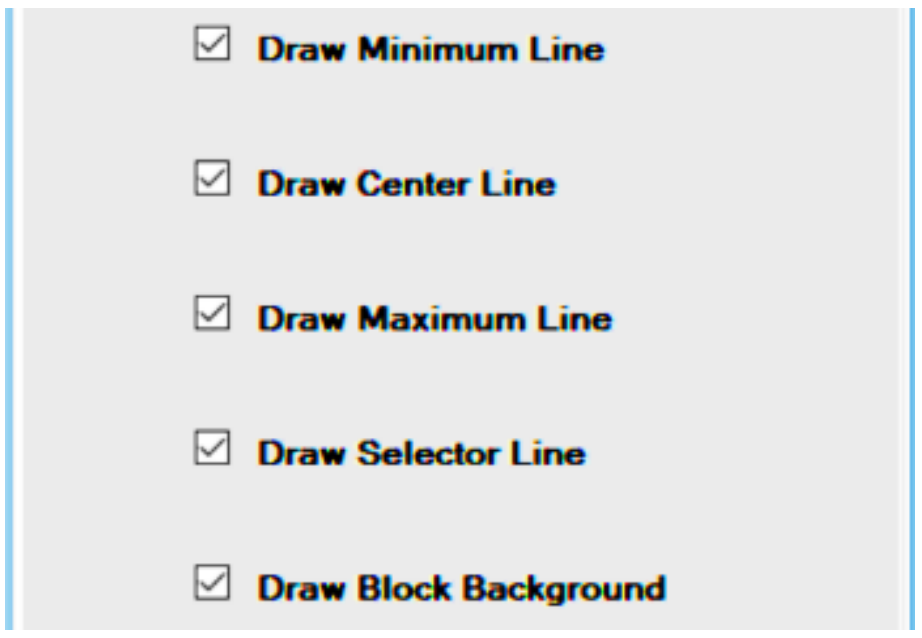

Fig. 22. Hyperblock selection utilities.

**Coloring lines in different colors**. Coloring each line in a different color allows distinguishing individual lines on all dataset and distinguishing different lines in subsets such as classes, hypercubes, and hyperblocks. Fig. 23 shows coloring every line a different color by using two coloring algorithms C2 and C3.

**Other features** that have been demonstrated above include showing a *histogram* of each coordinate to see where points overlap, *merging hyperblocks* to increase visual discovery generalization and automatically generating a *linguistic description* of a hyperblock and display it to the user.

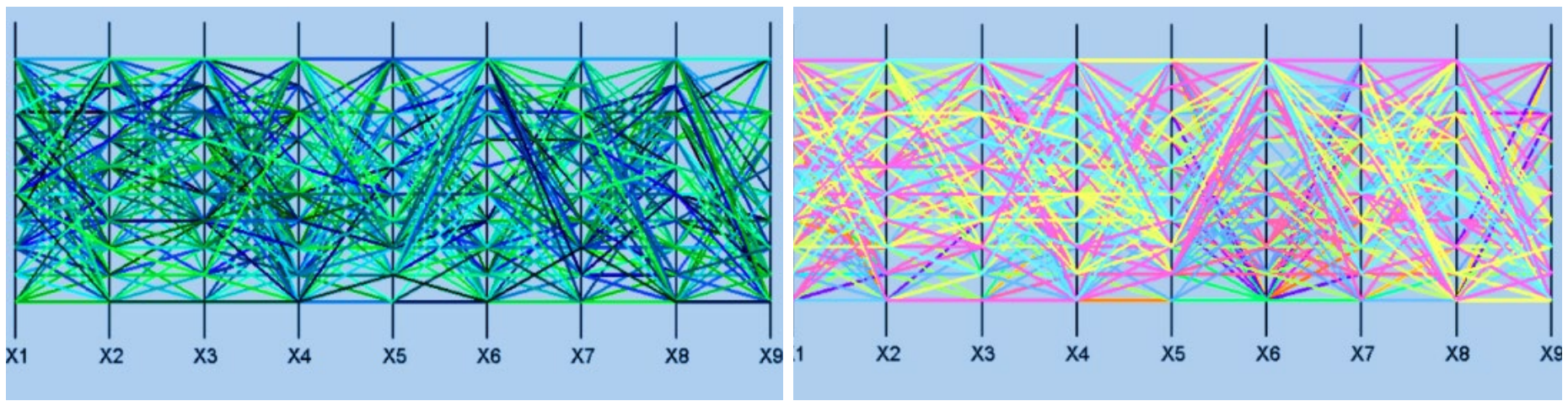

(a) (b)

Fig. 23. Individual lines colored by coloring algorithm C2 (a) and C3 (b).

The listed features implemented in VisCanvas 2.0 complement and expand features that were already implemented in VisCanvas 1.0 [12] making parallel coordinates adjustable. It includes abilities such as shifting coordinates up and down making the line of a given n-D point a straight horizontal line and reordering coordinates so that a selected line can grow or decrease in its values.

## 4.2.Missing Value Visualization

Commonly many datasets have a huge number of missing values (empty cells). Traditional approaches range from removing data with empty cells, to filling the missing values with some approximations computed automatically or conducted manually. Removing empty cells dramatically decreases the available data. In many datasets it can completely wipe out all data because by excluding the whole row or column with a single empty cell we are not only removing that single cell but all data in the respective row or column in the data table.

The missing values approximation is often risky because it can be incorrect. It also can be very time consuming if conducted manually. Therefore, it is necessary to develop a method which will overcome these difficulties, From the visualization viewpoint the major difficulty is that we cannot visualize attributes with missing values in parallel coordinates and in other general line coordinates too.

Below we present the **Empty Cell Visualization** method denoted as **ECV** method. Fig. 24 shows a typical dataset with missing values. The VisCanvas program analyzes such data files. If the value in the cell is not a number, but a text, then it is extracted and put as a label below the parallel coordinates plot as shown in Fig. 25.

If the cell is empty the word "Empty" is used as a label. See Fig. 25. In Figs 24 and 25 the labels include: n/c (not collected), did not record, Empty, ?, in other place, n /a (not applicable). Respectively, Fig. 25 visualizes that three blue cases have first two of these missing values in SL attribute. Other cases have different missing values in other attributes. Other datasets can have different terms and those terms are automatically extracted and put to the visualization. Thus, ECV method allows **"lossless" visual representation of missing values** preserving types of missing values.

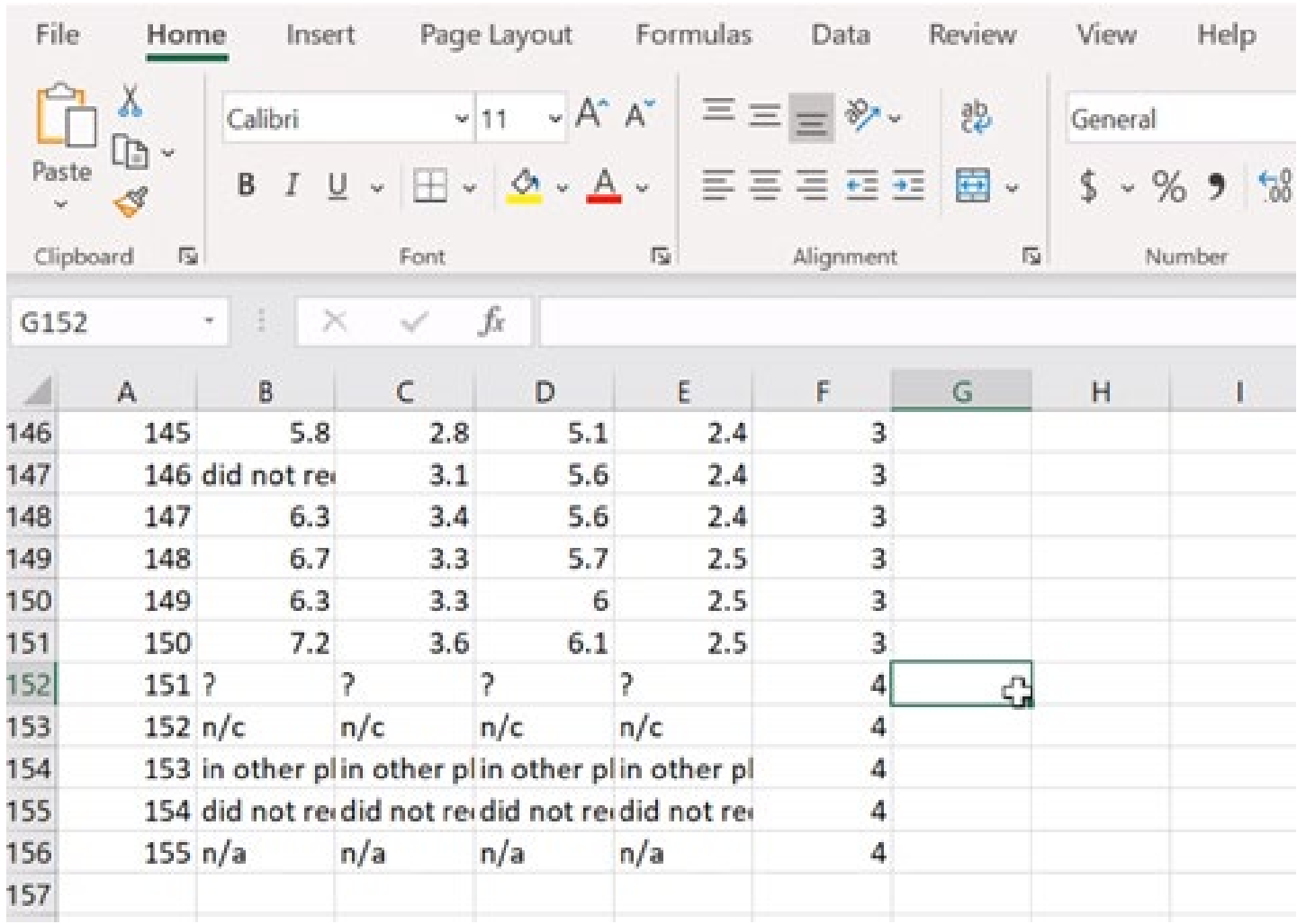

| | A | B | C | D | E | F | G | H | I |
|---|---|---|---|---|---|---|---|---|---|
| 146 | 145 | 5.8 | 2.8 | 5.1 | 2.4 | 3 | | | |
| 147 | 146 | did not re | 3.1 | 5.6 | 2.4 | 3 | | | |
| 148 | 147 | 6.3 | 3.4 | 5.6 | 2.4 | 3 | | | |
| 149 | 148 | 6.7 | 3.3 | 5.7 | 2.5 | 3 | | | |
| 150 | 149 | 6.3 | 3.3 | 6 | 2.5 | 3 | | | |
| 151 | 150 | 7.2 | 3.6 | 6.1 | 2.5 | 3 | | | |
| 152 | 151 | ? | ? | ? | ? | 4 | | | |
| 153 | 152 | n/c | n/c | n/c | n/c | 4 | | | |
| 154 | 153 | in other pl | in other pl | in other pl | in other pl | 4 | | | |
| 155 | 154 | did not re | did not re | did not re | did not re | 4 | | | |
| 156 | 155 | n/a | n/a | n/a | n/a | 4 | | | |
| 157 | | | | | | | | | |

Figure 24. Example of the dataset with missing values of different types.

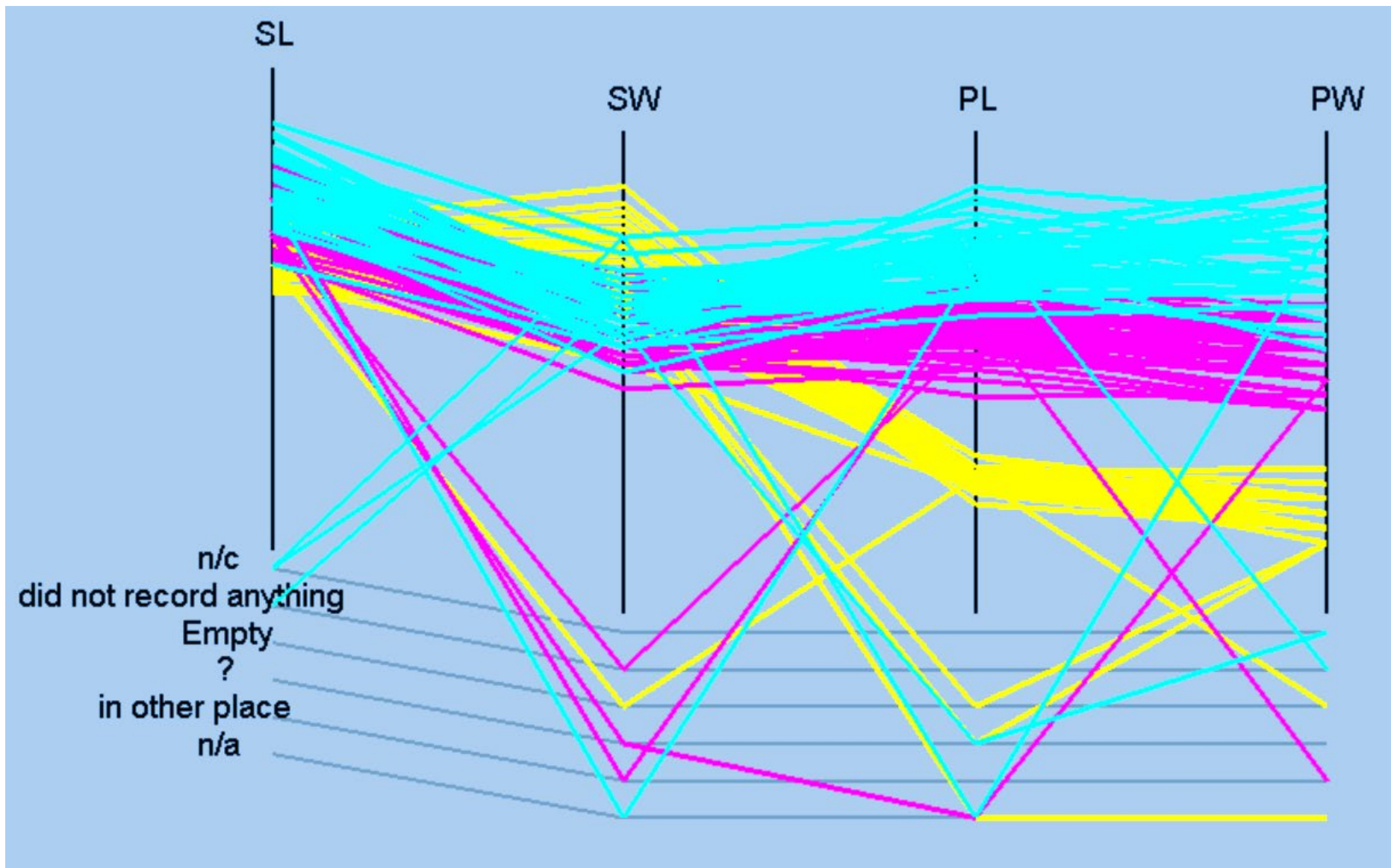

Fig. 25. Missing values visualized in Adjustable Parallel Coordinates.

In addition, Fig. 25 illustrates abilities of VisCanvas 2.0 to support **Adjustable Parallel Coordinates (APC)**, where each attribute can be drugged up and down to make any individual line straight.

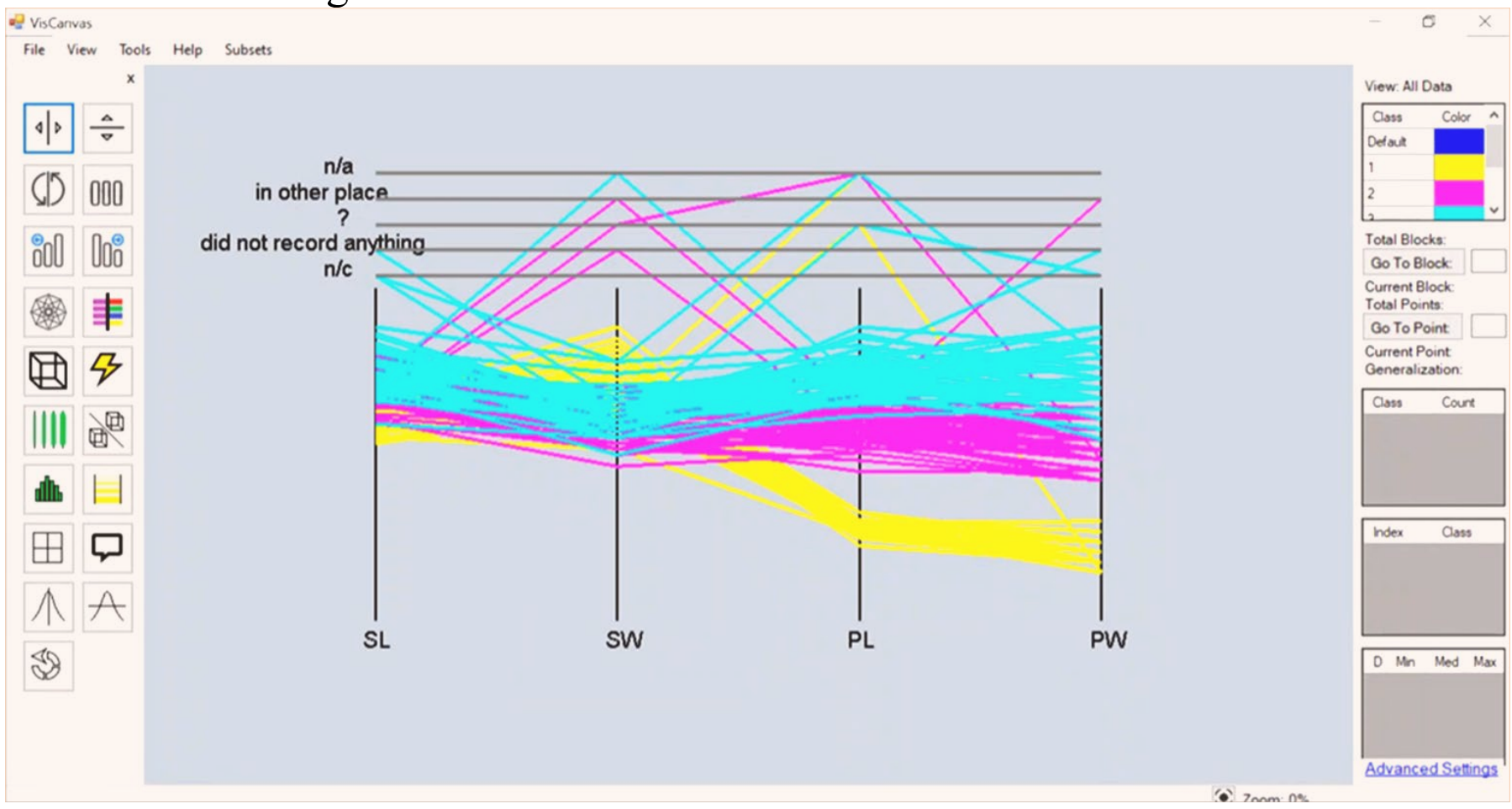

Fig. 26. Visualization of 4-dimentisonal data of 3 classes with markers of unknow values on the top.

This simplifies the comparison of that line with all other. See also Figs. 1, 2 in section 1. This line can be a center line of the HB or one of its edges. The perceptual simplification follows from the fact that straight horizontal lines are preattentive as it is established in the perception studies. Fig. 26 shows another example of visualization of a dataset with missing values with markers of missing values on the top. The benefit of the ECV approach is a dramatic expansion of abilities to visualize data with missing values, which was impossible before implementing in VisCanvas 2.0. This opens an opportunity to develop VisCanvas for deep analysis of data with missing values.

## 4.3.Dealing with large datasets

One of the significant issues is abilities interactively operate with the large data sets in parallel coordinates in VisCanvas 2.0. Specifically, we tested abilities of doing this in CWU testbed. We generated simulated 250,000 4-dimensional records. At the beginning rendering these data was very slow. Interactive manipulation like moving coordinates up/down was also very slow. Later, with modification of the code with a more efficient looping sequence we were able to do this instantly. See Figs. 26-27 show the result of manipulation with 250,000 records.

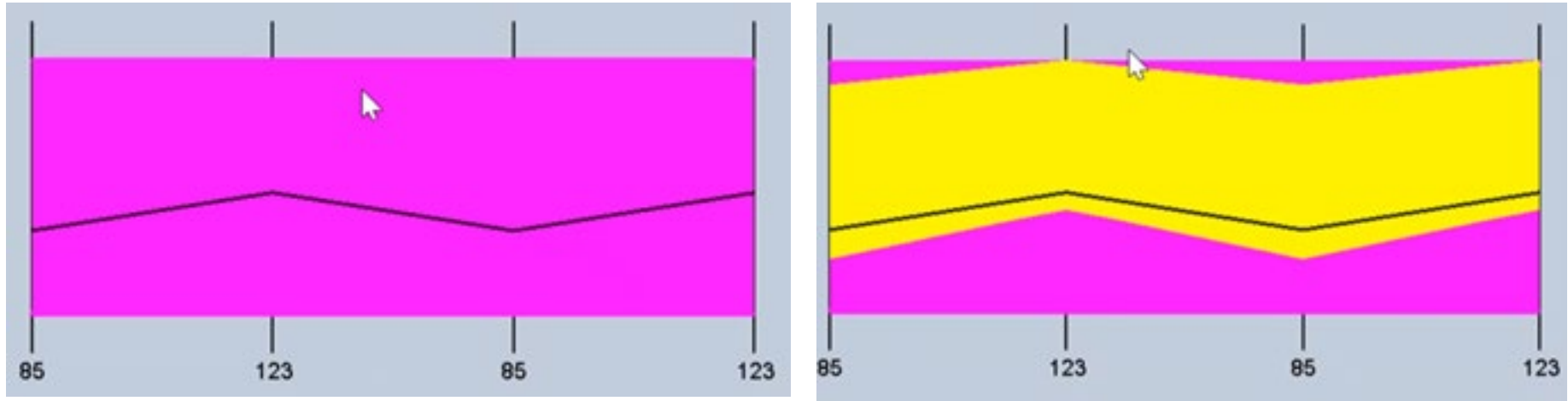

(a) Magenta class on the top.. (b) yellow class on the top.

Figure 27. 250,000 4-D records with a selected case (black)

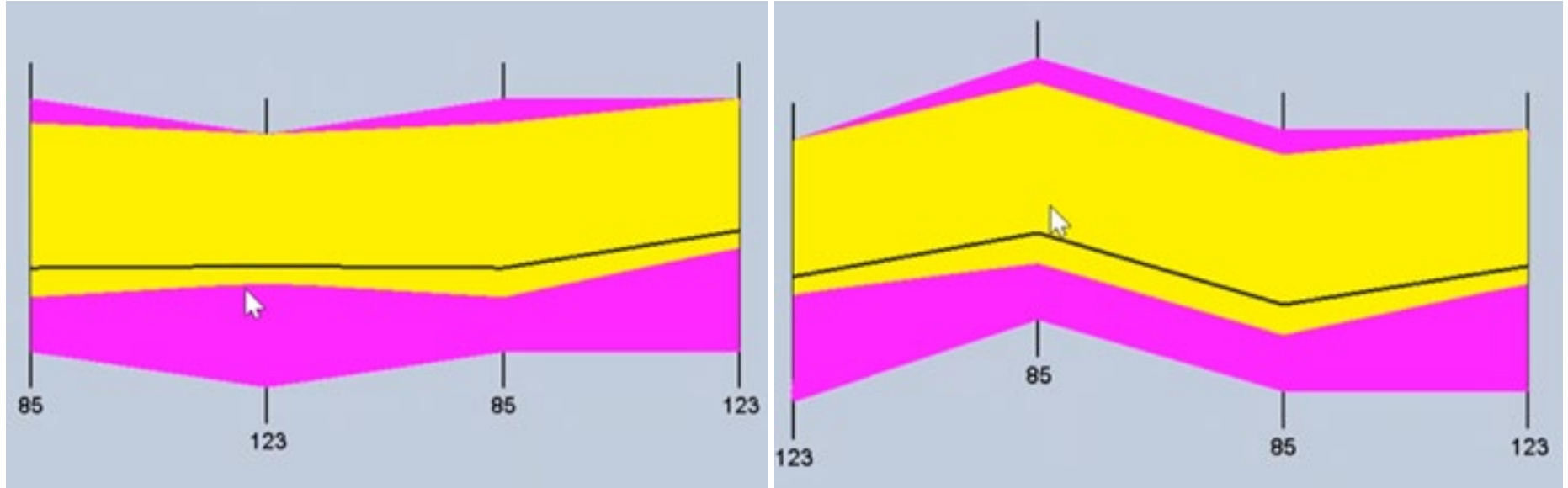

(a) Second coordinate shifted down. (b) Second coordinate shifted up.

Fig. 27. Real-time Interaction with 250,000 4-D records with shifting coordinates up/down

## 6. DISCUSSION, GENERALIZATION,CONCLUSION AND FUTURE WORK

**Hyperblocks in parallel coordinates**. This chapter contributes to interpretable machine learning via visual knowledge discovery in parallel coordinates by visual pattern discovery, data and model classification visualization, dimension reduction and model simplification. This allows putting the end-user in the driver seat of model development. It is impressive that [17] traced parallel coordinates to 19th century [16] to visualize the US Census data. See Fig. 28 reproduced from [17].

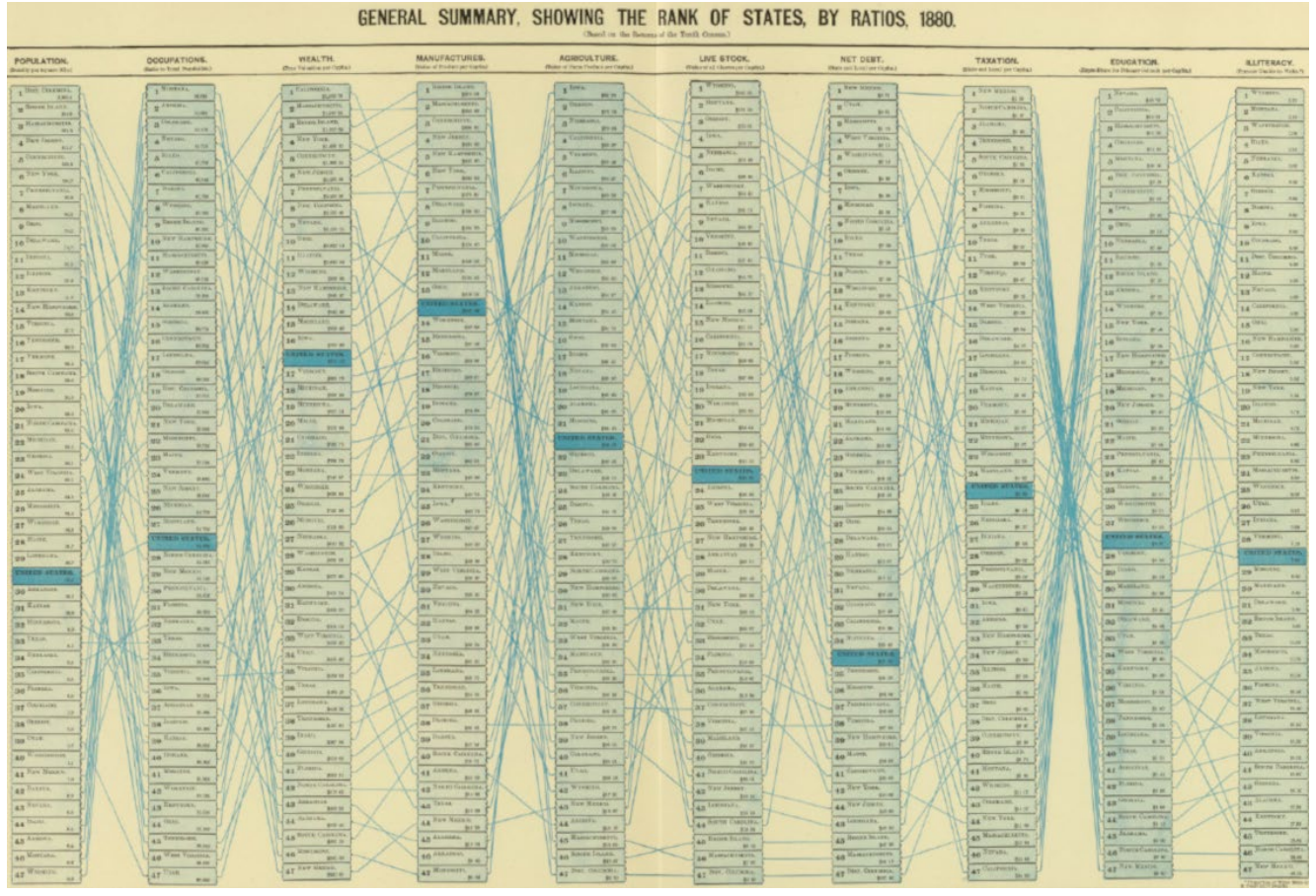

Fig. 28. Early parallel coordinates visualization [16] showing the United States ranked by population, wealth, livestock, net debt, and others. The blue observation represents the U.S. average.

Parallel coordinates were dormant and unknown for almost 100 years. Now parallel coordinates are actively used to explore n-D data. However, the use of parallel coordinates to improve supervised learning algorithms is still in the nascent stage.

The concepts of *hypercubes* and *hyperblocks* are used as the major concepts in this study to allow end-users to easily understand patterns on parallel coordinates. The base definitions of the hypercubes and hyperblocks are provided along with their very intuitive visualization in parallel coordinates. Then we described algorithms to discover HBs interactively and automatically in multiple HB settings: individual and multiple, overlapping, and non-overlapping, and in combination of hyperblocks with linguistic description of visual patterns.

This paper discussed challenges of classification algorithms based on hyperblocks and proposed the Hyper algorithm for classification with mixed and pure hyperblocks. It is shown that Hyper models generalize decision trees. The Hyper algorithm was tested on the benchmark Wisconsin Breast Cancer data. It allowed discovering pure and mixed HBs with all these data and then with 10-fold cross validation. We established a link between hyperblocks, dimension reduction and visualization of lower dimensional HBs.

**Hyperblocks in other general line coordinates**. The HBs are simple and interpretable units in supervised machine learning. It creates an opportunity to use them in other General Line Coordinates beyond parallel coordinates. Below Figs. 28, 29 shows an example of a two hyperblocks (blue and red) in the Shifted Paired Coordinates (SPC) and Collocated Paired Coordinates (CPC) defined in [6]. Advantage of the visualization in SPC is that these blue and red rectangles do not overlap in contrast with them in CPC and parallel coordinates (see Fig. 29). SPC and CPC require only three simple rectangles for each HB but in parallel coordinates requires five more complex 4-gones.

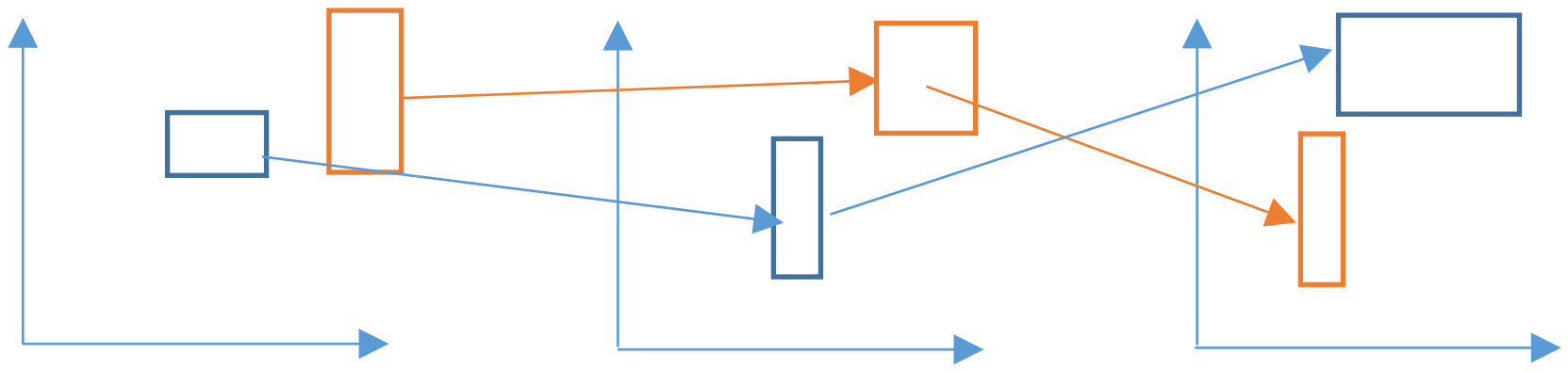

Fig. 28. Red and blue hyperblocks in Shifted Paired Coordinates as connected rectangles.

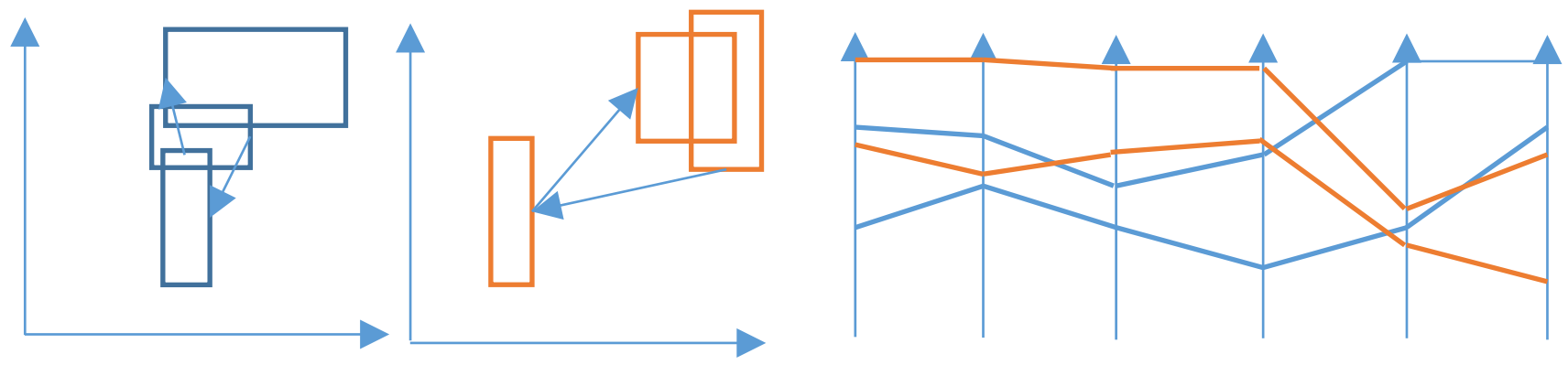

(a) Blue HB in CPC. (b) Red HB in CPC. (c) Red and blue HBs in parallel coordinates.

Fig. 29. Red and blue hyperblocks in Collocated Paired and Parallel Coordinates.

Major benefits of hyperblock technology and Hyper algorithm are the abilities to discover and observe hyperblocks by end-users. It includes observing all HBs together side by side (see. Fig. 12) making patterns visible for all classes. Also, it allows detailed visual analysis of individual HBs of individual classes. HBs are well understood by the end-users in the original attributes with the ability to cut down unimportant attributes. Another advantage of HBs relative to DTs is abilities to avoid overgeneralization of data. DTs set up thresholds in each node that can be far away from actual data of classes. Fig. 14 for HBs and Fig. A1 in Appendix for DT illustrate this situation.

**Empty Cell Visualization.** The paper also presented features of VisCanvas 2.0 software that implement Hyper algorithms and other HB functionality including Empty Cell Visualization method and dealing with large datasets. The conducted studies have shown that we can successfully deal with empty cells and large datasets. It also opens an opportunity to generalize the empty cell visualization method to other General Line Coordinates, like Shifted Paired Coordinates. The idea is that empty cell marker can be located at the beginning and/or end of each line coordinate, as it is shown in Fig. 30 for Shifted Paired Coordinates. Future studies include maturing the algorithms presented to the full scope of machine learning in 2-D parallel coordinates instead of only analytically in n-D. The VisCanvas software system will be expanded to support larger datasets using GPU and multithreading.

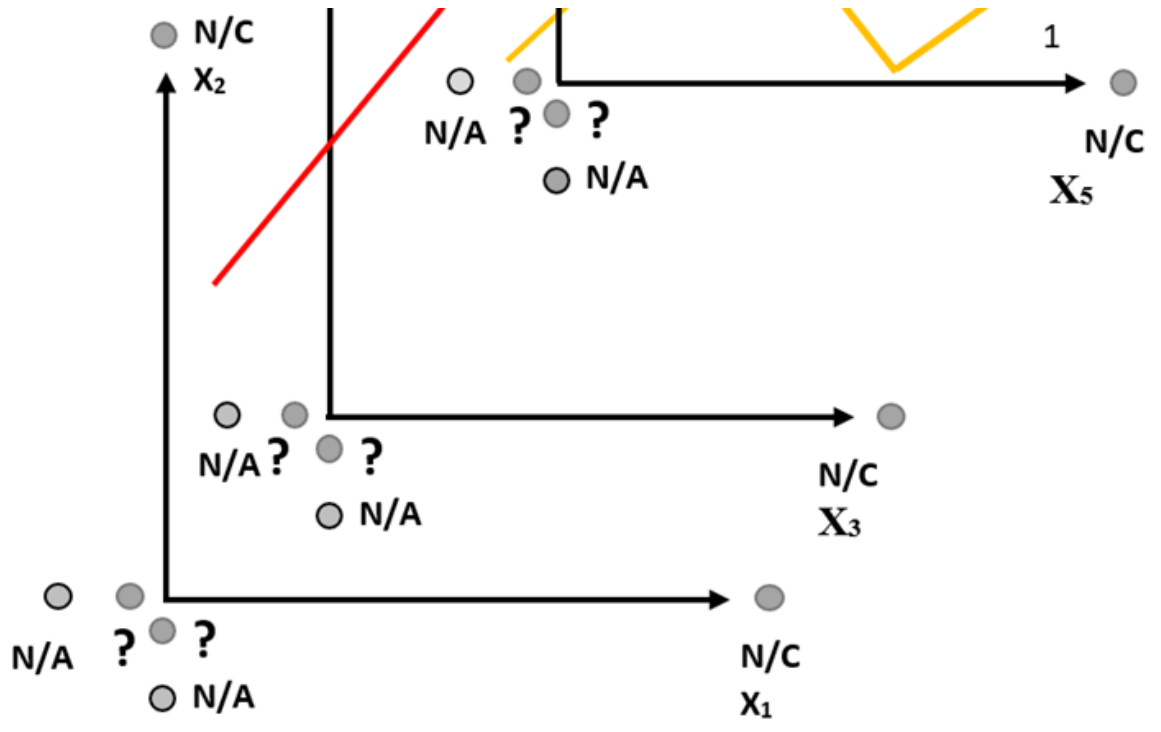

Fig. 30. Visualization of 6-D data with empty cells in Shifted Paired Coordinates.

Another aspect of the future work is usability studies. Multiple prior studies with parallel coordinates already had shown that users can easily capture simple pattern in parallel coordinates. Hyperblocks belong to the category of simple patterns. This is a major reason why we concentrate this work on hyperblocks. Therefore, a traditional usability study where hyperblocks will be shown to the users with asking if a user considers them simple or not will not be much informative. We plan a more informative usability study where a domain expert without advanced knowledge of machine learning will discover and interpret pure enough hyperblocks using VisCanvas 2.0 tool.

Running Hyper algorithm of large datasets requires significant runtime. The future work will be on parallelization of the computational processes, efficient data buffering, using multi-threading, powerful GPU and supercomputer resources. It will allow for operations to take place in the background while the user is still able to interact with the UI. The challenges involved are identifying tasks, balance, data splitting, and data dependency. Both data and task parallelization are future challenges. To help shortening the run time big datasets can be split into smaller sets handled by multiple threads while the work done on them can also be parallelized. The use of thread pools, where threads are already instantiated and waiting to be requested, will be another way to explore and deal with the big data.

## APPENDIX

Table A1. Branches of ID3 Decision tree for the WBC data (class 2 - B, class 4- M).

<table>
<tr>
<td rowspan="2">|--- x2 <= 2.50<br>| |--- x6 <= 5.50<br>| | |--- x8 <= 9.00<br>| | | |--- x1 <= 7.50<br>| | | | |--- class: 2<br>| | | |--- x1 > 7.50<br>| | | | |--- x6 <= 2.00<br>| | | | | |--- class: 2<br>| | | | |--- x6 > 2.00<br>| | | | | |--- class: 4<br>| | |--- x8 > 9.00<br>| | | |--- class: 4<br>| |--- x6 > 5.50<br>| | |--- x1 <= 2.50<br>| | | |--- class: 2<br>| | |--- x1 > 2.50<br>| | | |--- class: 4</td>
<td>|--- x2 > 2.50<br>| |--- x3 <= 2.50<br>| | |--- x1 <= 5.50<br>| | | |--- class: 2<br>| | |--- x1 > 5.50<br>| | | |--- class: 4<br>| |--- x3 > 2.50<br>| | |--- x2 <= 4.50<br>| | | |--- x6 <= 2.50<br>| | | | |--- x5 <= 3.50<br>| | | | | |--- class: 2<br>| | | | |--- x5 > 3.50<br>| | | | | |--- x5 <= 7.00<br>| | | | | | |--- class: 4<br>| | | | | |--- x5 > 7.00<br>| | | | | | |--- class: 2</td>
<td rowspan="2">|--- x2 > 2.50<br>| |--- x3 <= 2.50<br>| | |--- x2 > 4.50<br>| | | |--- x4 <= 1.50<br>| | | | |--- x1 <= 7.00<br>| | | | | |--- x9 <= 2.00<br>| | | | | | |--- class: 2<br>| | | | | |--- x9 > 2.00<br>| | | | | | |--- class: 4<br>| | | | |--- x1 > 7.00<br>| | | | | |--- class: 4<br>| | | |--- x4 > 1.50<br>| | | | |--- x8 <= 2.50<br>| | | | | |--- x8 <= 1.50<br>| | | | | | |--- class: 4<br>| | | | | |--- x8 > 1.50<br>| | | | | | |--- x6 <= 9.00<br>| | | | | | | |--- class: 2<br>| | | | | | |--- x6 > 9.00<br>| | | | | | | |--- class: 4<br>| | | | |--- x8 > 2.50<br>| | | | | |--- class: 4</td>
<td rowspan="2">|--- x2 > 2.50<br>| |--- x3 > 2.50<br>| | |--- x2 <= 4.50<br>| | | |--- x6 > 2.50<br>| | | | |--- x1 <= 6.50<br>| | | | | |--- x1 <= 5.50<br>| | | | | | |--- x6 <= 6.00<br>| | | | | | | |--- x4 <= 3.50<br>| | | | | | | | |--- x3 <= 3.50<br>| | | | | | | | | |--- class: 4<br>| | | | | | | | |--- x3 > 3.50<br>| | | | | | | | | |--- class: 2<br>| | | | | | | |--- x4 > 3.50<br>| | | | | | | | |--- class: 4<br>| | | | | | |--- x6 > 6.00<br>| | | | | | | |--- x7 <= 3.50<br>| | | | | | | | |--- x5 <= 6.50<br>| | | | | | | | | |--- class: 4<br>| | | | | | | | |--- x5 > 6.50<br>| | | | | | | | | |--- class: 2<br>| | | | | | | |--- x7 > 3.50<br>| | | | | | | | |--- class: 4<br>| | | | | |--- x1 > 5.50<br>| | | | | | |--- class: 2</td>
</tr>
<tr>
<td>|--- x2 > 2.50<br>| |--- x3 > 2.50<br>| | |--- x2 <= 4.50<br>| | | |--- x6 > 2.50<br>| | | | |--- x1 > 6.50<br>| | | | | |--- x8 <= 7.00<br>| | | | | | |--- class: 4<br>| | | | | |--- x8 > 7.00<br>| | | | | | |--- x8 <= 8.50<br>| | | | | | | |--- x1 <= 7.50<br>| | | | | | | | |--- class: 4<br>| | | | | | | |--- x1 > 7.50<br>| | | | | | | | |--- class: 2<br>| | | | | | |--- x8 > 8.50<br>| | | | | | | |--- class: 4</td>
</tr>
</table>

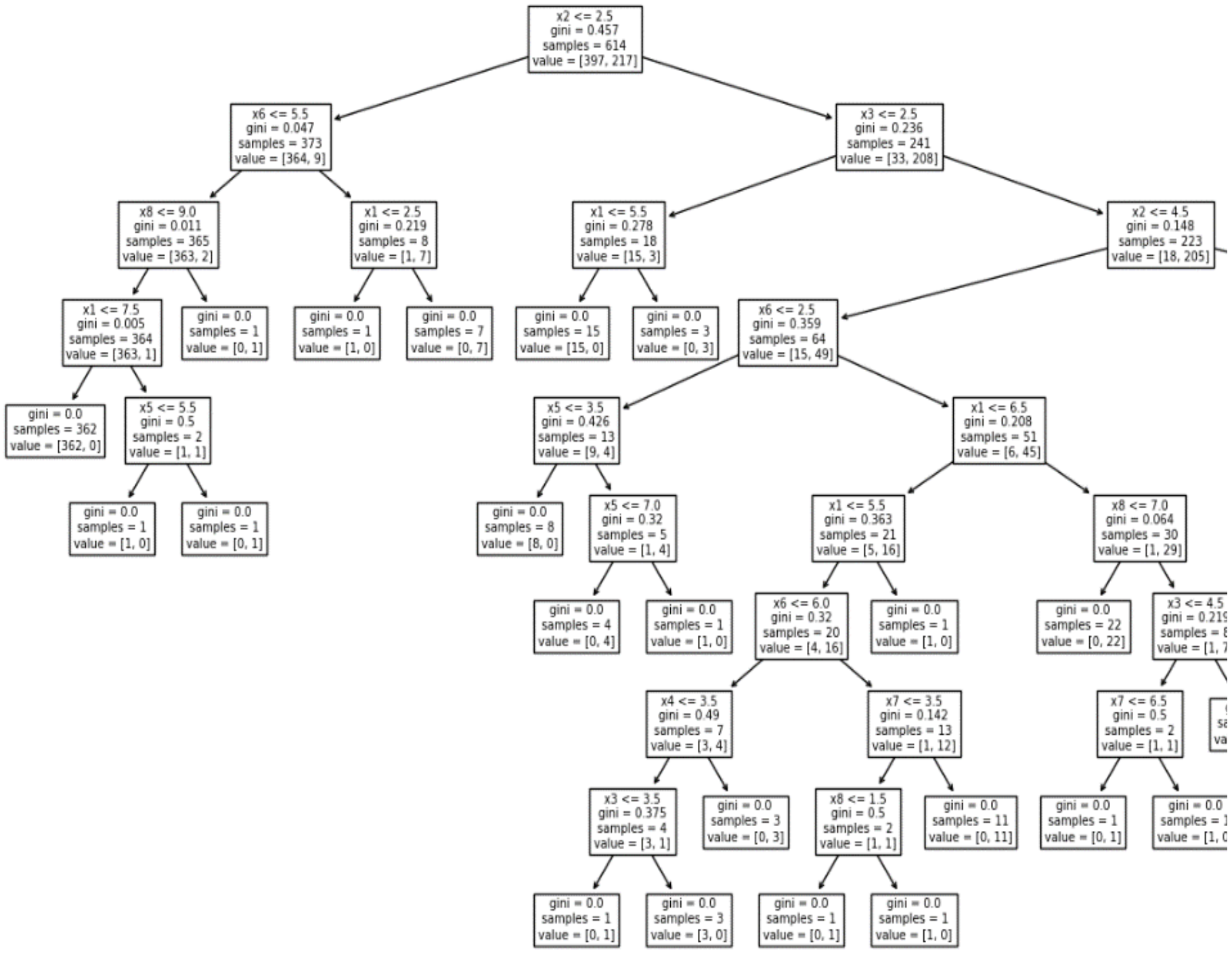

Fig. A1. ID3 Decision tree for WBC data with multiple “mini” terminal nodes.